\documentclass{article}

\usepackage{arxiv}

\usepackage[utf8]{inputenc} 
\usepackage[T1]{fontenc}    
\usepackage{hyperref}       
\usepackage{url}            
\usepackage{booktabs}       
\usepackage{amsfonts}       
\usepackage{nicefrac}       
\usepackage{microtype}      

\usepackage{xspace}
\usepackage{xcolor}
\usepackage{amstext}
\usepackage{amsfonts}
\usepackage{mathtools}
\usepackage{multirow}

\newcommand{\newfixme}[1]{    #1   }
\newcommand{\fixme}[1]{    #1   }
\newcommand{\revision}[1]{  #1  }

\newcommand\blfootnote[1]{%
  \begingroup
  \renewcommand\thefootnote{}\footnote{#1}%
  \addtocounter{footnote}{-1}%
  \endgroup
}

\title{DeepFlash: Turning a Flash Selfie into a Studio Portrait}

\author{
  Nicola~Capece\\
  University of Basilicata\\
  Potenza, Italy \\
  \texttt{nicola.capece@unibas.it} \\
   \And
 Francesco~Banterle \\
  ISTI-CNR\\
  Pisa, Italy \\
  \texttt{francesco.banterle@isti.cnr.it} \\
  \And
 Paolo Cignoni \\
  ISTI-CNR\\
  Pisa, Italy \\
  \texttt{Paolo.Cignoni@isti.cnr.it} \\
     \And
 Fabio Ganovelli \\
  ISTI-CNR\\
  Pisa, Italy \\
  \texttt{fabio.ganovelli@isti.cnr.it} \\
     \And
 Roberto Scopigno \\
  ISTI-CNR\\
  Pisa, Italy \\
  \texttt{roberto.scopigno@isti.cnr.it} \\
     \And
 Ugo Erra \\
University of Basilicata\\
  Potenza, Italy \\
  \texttt{ugo.erra@unibas.it}
}

\begin{document}
\maketitle

\begin{abstract}
We present a method for turning a flash selfie taken with a smartphone into a photograph as if it was taken in a studio setting with uniform lighting.
Our method uses a convolutional neural network trained on a set of pairs of photographs acquired in an ad-hoc acquisition campaign. Each pair consists of one photograph of a subject's face taken with the camera flash enabled and another one of the same subject in the same pose illuminated using a photographic studio-lighting setup. 
We show how our method can amend defects introduced by a close-up camera flash, such as specular highlights, shadows, skin shine, and flattened images\blfootnote{Published as a journal paper at Signal Processing: Image Communication. https://doi.org/10.1016/j.image.2019.05.013}.
\end{abstract}


\keywords{
Image enhancement \and Machine learning algorithms \and Deep Learning \and Computational photography \and Image processing}

\section{Introduction}

\begin{figure}
 \includegraphics[width=\linewidth]{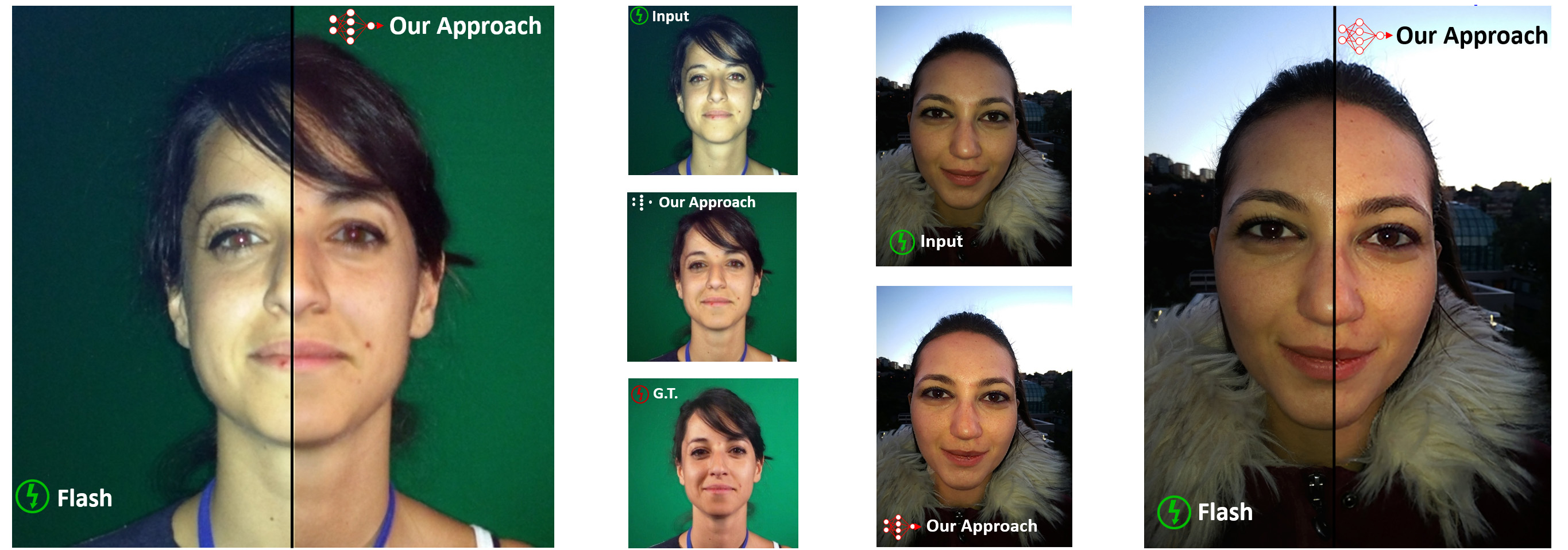}
 \centering
  \caption{
  Two examples from our results. The split images show a comparison between the input and the output of our algorithm. In the central column the input, output, and ground truth images.}
\label{fig:teaser}
\end{figure}

With the steady improvement of built-in digital cameras, pictures taken on smartphones and computer tablets are becoming increasingly predominant on the Internet, even on web-based services dedicated to quality photography such as Flickr, 500px, and Instagram.
Although smartphones can take pictures that are comparable to those of digital reflex cameras in favorable lighting conditions, they perform poorly in low light conditions.
This is mainly due to the size of their sensors, which is a constraint difficult to overcome within the small room on a smartphone.
Therefore, taking pictures in low light often triggers the camera flash, which is typically a low-power LED flash mounted side by side with the camera lens that \revision{produces several artifacts.}


\noindent One of the most common type of photograph taken with a smartphone is the so-called selfie, which is a picture of one's face taken by holding the phone in the  hand or by using a ``selfie stick''. Low-light flash photographs and selfies are an unfavorable combination that produces images with specular highlights, sharp shadows, and flat and unnatural skin tones. \newfixme{Therefore, researchers have recently started to develop several correction techniques: re-lighting and enhancement of images with non-uniform lighting  \cite{bychkovsky2011learning,wang2013naturalness}, some of them have focused mainly on the images of faces \cite{shih2014style,wang2009face,wen2003face,shu2018portrait}.}
In this paper, we explore the possibility of taking smartphone flash selfies and employing a convolutional neural network (CNN) to turn  them into studio portraits.
Researchers have extensively used CNNs to improve pictures, for example for creating high dynamic range images from single exposure~\cite{EKDMU17}, colorization~\cite{Iizuka16}, super-resolution~\cite{Ledig17}, and so on, as will be discussed in Section~\ref{sec:DLCP}. However, our problem is especially challenging for at least three reasons. Firstly, it concerns an effect that has both local and global discriminant features such as highlight and skin tone, respectively. Secondly, we want to imitate a process that humans are very good at performing; i.e., to picture what an image would look like if flash was not used. Finally, both previous points apply to the domain of human faces, on which humans are extremely good to spot any kind of inconsistencies.
We leverage the fact that, by their nature, smartphone flash selfies share many common traits and make a fairly well-defined subdomain of photographs: they are front or three-quarter single-face portraits, taken from less than one meter away with a single flash located with the camera lenses.
Our approach consists of training a CNN with a series of pairs of portraits, where one is taken with the smartphone flash and one with photographic studio illumination. The two photographs of the same pair are taken as simultaneously as possible, so that the pose of the subject is the same. 
\revision{Although the flash no-flash problem applies to a wider application domain, the collection of images useful for this aim is still a challenge today, considering the huge amount of data required \cite{deng2009imagenet} (i.e., the order of hundreds of thousand). As explained by Sun et al. \cite{sun2017revisiting},  focusing on a specific category for a deep learning algorithm is very important, whereas adding random categories could reduce the performance. For this reason, the choice of the restricted dataset and task to improve the only selfie leads to a more specialized deep learning algorithm and results in higher performance. 
Moreover, one of our contributions was to build a specific dataset, which size represents a lower bound useful to train CNN to perform this task.}

The remainder of this paper is structured as follows: In section~\ref{sec:related}, we review the literature most relevant to our problem. In section~\ref{sec:ourApproach}, we introduce the network we designed and the way it is used to solve our problem. In section~\ref{dataset_creation}, we describe the experimental  setup used to create the training dataset. In section~\ref{sec:results}, we show  and discuss the results obtained with our method, in section~\ref{sec:limitations}, we comment on its limitations and finally we draw our conclusion and address future work in section~\ref{sec:conclusion}.

\section{Related Work}
\label{sec:related}
In this section, we review works related to our work and photography per se; that is, without a focus on acquisition for computer vision/graphics tasks (e.g., color acquisition for three-dimensional (3D) meshes).

\subsection{Flash Photography}
In two contemporary works, Petschnigg et al. \cite{Petschnigg+2004} and Eisemann and Durand \cite{Eisemann+2004} proposed the idea of using a flash photograph with low ISO (i.e., low noise) to transfer the ambiance of the available lighting into a non-flash photograph of the same subjects/scene, to reduce noise. This is because non-flash photographs taken in dark environments suffer from high noise to avoid blur. Other works\cite{Agrawal+2005} have developed this idea further by removing over/under-illumination at a given flash intensity, reflections, highlights, and attenuation over depth.

\subsection{Deep Learning}
Deep learning (DL) \cite{LeCun+2015} is a framework inspired by biology that enables computational models (made of nonlinear projection) to learn high-level representation from data, by \revision{back-propagating errors~\cite{rumelhart1986learning}} with respect to the model parameters. The most important advantage of DL is that it avoids feature engineering that is automatically inferred from raw data.
One class of DL architecture of particular interest is the CNN class, namely, DL models that mimic how the visual cortex works.
Krizhevsky et al.'s work \cite{Krizhevsky+2012} on CNNs  has sparked a renaissance in the field of DL. Furthermore, this work has pushed forward the state of the art in many imaging tasks. This has been made possible because of three key factors: the increasing computing power of modern graphics processing units (GPUs), the availability of huge datasets, and novel and powerful algorithms and frameworks for CNNs.
To improve their learning invariance, CNNs are typically trained on augmented data\cite{Dosovitskiy+2014}; as will be shown later, data augmentation was found to be essential in our experimentation.

\subsection{Deep Learning and Computational Photography}\label{sec:DLCP}
Recently, CNNs have been applied to numerous computer vision, imaging, and computer graphics tasks \newfixme{\cite{shu2017neural}\cite{sfsnetSengupta18}\cite{ignatov2017dslr}\cite{chen2018deep}\cite{hu2018exposure}\cite{zhu2017unpaired}}. Furthermore, they are becoming extremely popular, and novel architectures and algorithms are continually popping up overnight.

In this section, we review works with similar aims to ours, namely, starting from a single image and attempting to improve its quality or aesthetics.

To improve the editing of selfies, Shen et al. \cite{Shen+2016} extended the FCN-8s framework \cite{Shelhamer+2017} to automatically segment portraits. This allows a user to automatically edit/augment portraits. For example, users can change the background, stylize the selfie, enhance the depth-of-field, etc. Zhang et al. \cite{Zhang+2018S} proposed an end-to-end learning approach for single-image reflection separation with perceptual losses and a customized exclusion loss. Their method can be used to remove or reduce unwanted reflections in pictures. 

Eilertsen et al. \cite{EKDMU17} introduced a U-Net architecture \cite{ronneberger2015u} for expanding the dynamic range of low dynamic range images to obtain high dynamic range images.
Similarly, Chen et al. \cite{Chen+2018} showed that U-Nets can be used successfully to debayer images captured at low-light conditions and high ISO, which  typically exhibit noise. They extensively studied different approaches on real-world low light with real noise. For example, they tested a variety of architectures, loss functions (e.g., L1 (least absolute deviations), L2 (least square errors), and the structural similarity index (SSIM)), and color inputs.

\newfixme{Aksoy et al. ~\cite{flashambient} presented a large-scale collection of pairs of images with ambient light and flashlight of the same scene. The images were obtained with the aid of casual photographers by using their smartphone cameras, and consequently, the dataset covers a wide variety of scenes. The dataset was provided in order to be studied in future works for high-level tasks such as semantic segmentation or depth estimation. Unlike their dataset, whose objective is to provide matching between two images under uncontrolled lighting conditions, our dataset aims to change the lighting scheme by turning from flash lighting to a controlled photograph studio light.}

To the best of our knowledge, there have not been proposed approaches for removing flash artifacts such as hard shadows and highlights from selfies as the work presented in this paper.

\section{Our Approach}
\label{sec:ourApproach}

We designed a regression model targeted to the restricted domain of human faces. We adopted a supervised approach where a CNN is trained by feeding pair of flash and no flash portraits. Although the main idea is straightforward, there are several details that need to be addressed in the design of the network, the training procedure, \revision{the way the problem is encoded in the network, and finally the loss function.} All these aspects are discussed in the following sections.

\begin{figure}[!ht]  
	\centering
	\includegraphics[width=\linewidth]{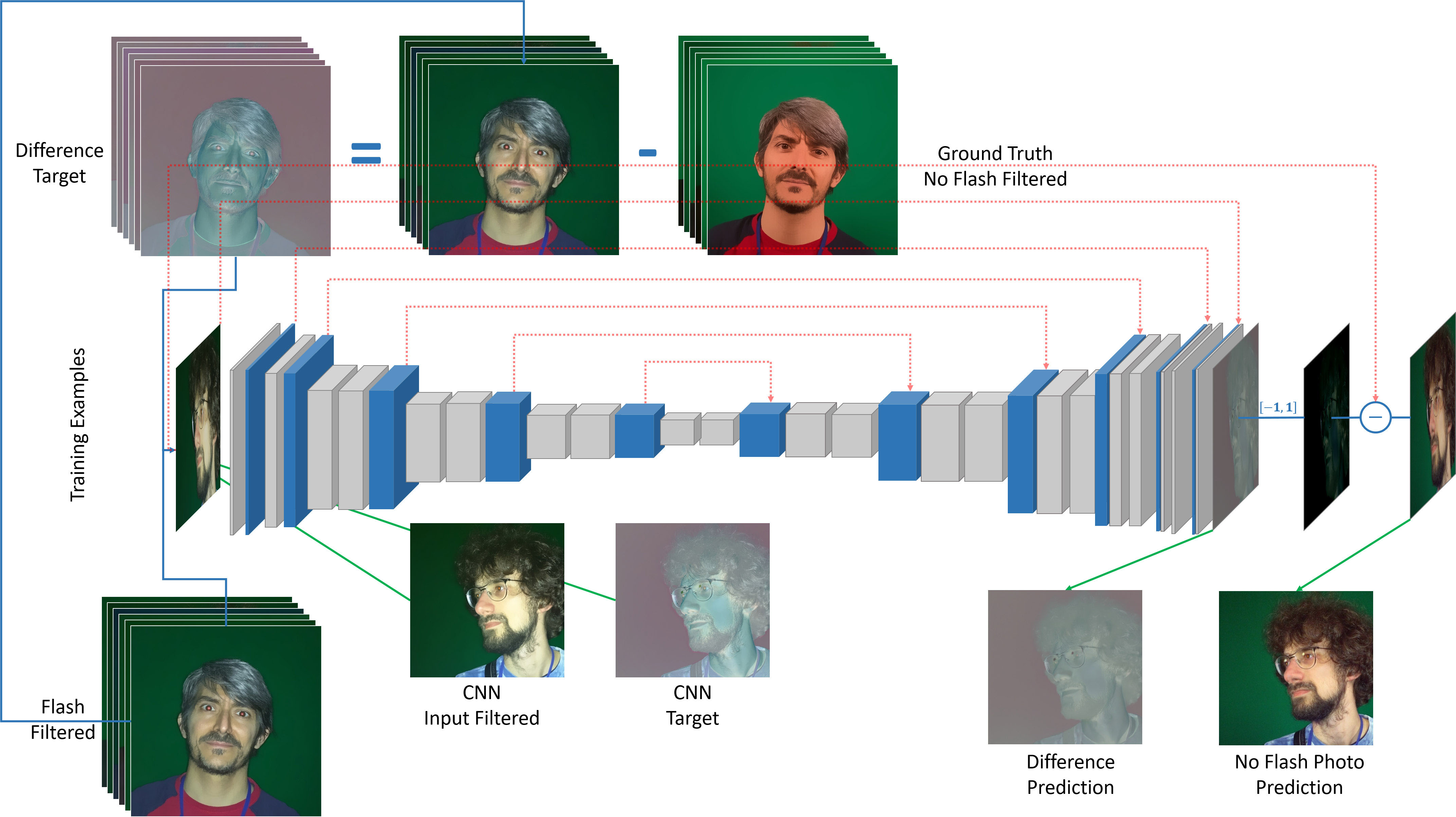}
	\caption{Our neural network architecture for transforming a flash image into a non-flash image. The first 13 blocks represent the VGG-16's convolutional layers, which perform the image encoding. The second part reconstructs the output image and has several convolutional and deconvolutional layers. From the blue blocks of the VGG-16, the shortcut connections start, which are linked with their counterparts in the decoder. The CNN input is an image taken with the smartphone flash, and the ground truth is an image taken using simulated ambient light, on both of which the bilateral filter is applied. The target image is the difference between the input and ground truth image, normalized in a range between 0 and 1. The network prediction is the searched difference, which is denormalized in the range $[-1,1]$ and then subtracted from the non-filtered input. The final output prediction is an image without flash highlights.}
\label{fig:ourApproach}
\end{figure}

\subsection{Deep Neural Network} 


Our CNN is an \fixme{encoder-decoder structured neural network}  that consists of two sub-networks: the first network takes as input a flash image and performs the encoding to create a deep feature map representation; the second network takes as input the encoder's output and recreates the image without the flash defects. The input image encoding is performed by the well-known Visual Geometry Group's VGG-16\cite{Simonyan14c} network. This is a  CNN widely used for object detection and classification tasks, achieving 92.7\% accuracy in results on 1000 classes of objects in ImageNet (a dataset of about 14 million images).

The network is composed of 16 layers, 13 of which are convolutional layers, and 3 are fully connected layers. For our purpose, we used only the convolutional layers, which are divided into five separate groups: the first two groups and the last three are composed, respectively, of two and three layers (Figure \ref{fig:ourApproach}). Each single CNN layer performs three operations in sequence: (i) A set of linear activations is produced using several parallel convolutions by a $3 \times 3$ kernel size, a stride value of 1, and a padding value set to ``SAME'' to preserve the spatial resolution between two convolutions; (ii) linear activation is performed through a nonlinear activation function called ReLU (Rectified Linear Unit)\cite{pmlr-v15-glorot11a}, which is defined as $ReLU(x) = max(0, x)$, and ensures that the gradients remain large and consistent; (iii) a max pooling \cite{Zhou-1988} operation takes the output of the convolutional layer as input and reduces its size to match the input size of the next layer through a downsampling operation. In particular, max pooling takes the largest value from a window moved over the output of the convolutional layer. In this way, the extracted values from the pooling operation do not vary \cite{Goodfellow-et-al-2016}, and the output is invariant to small translations of the input.

We developed the \fixme{decoder} component, which performs the decoding task, based on Eilertsen et al.'s approach\cite{EKDMU17}. In particular, the input of the network is the output from the last VGG-16 convolutional layer after a further convolution operation. As the change of parameters of the CNN in the training phase changes the output distribution of each layer, it is necessary to normalize this distribution to produce an output that will be valid as the input for the next layer. For this reason, batch normalization \cite{ioffe2015batch} is performed after each convolution, forcing the input of the activation function to have mean zero and unit variance.

After each batch normalization, the obtained tensor, called the activation tensor, crosses the next activation function, LeakyReLU \cite{Maas13rectifiernonlinearities}, which is a modified version of ReLU introducing a nonzero gradient for the next inputs. To obtain optimal CNN performance in the training phase, the slope parameter $\alpha$ is set to 0.2 \cite{Xu2015EmpiricalEO}.

The main features of the decoder layers are operations such as convolutions, batch normalization, deconvolutions, and concatenations. 
Because our network is composed of many layers, to avoid the vanishing gradient problem \cite{Hochreiter01gradientflow} studied by He et al.\cite{He-2016}, we used an approach based on a residual learning network. This problem concerns the weight update in the backpropagation phase, which is proportional to the gradient of the error function compared with the weight that has to be updated. Progressing in the backpropagation phase and inverse crossing the network to update the weight may mean that the gradient is so small as to make inefficient updates on the weights belonging to the first deep neural network layers, and consequently, their training is stopped. A simple way to solve this problem is to use a block division of the VGG-16 and concatenate in depth each block's output with its counterpart in the decoder through a link called a shortcut connection. For this reason, we use concatenations layers.

In the image recognition task, the residual learning network is based on the assumption that the network has to learn a residual function. If we consider $H(x)$ to be the function that a layers group has to learn, and $F(x)$ to be the function that a layers group is able to approximate, whether at this function is concatenated to the group's input, it will obtain a function in the form $F(x) = H(x) - x$ and a residual function in the form $H(x) = F(x) + x$. In this way, the contiguous layers that belong to the group learn the residual function, and their input will be concatenated to their own output. A similar approach is often used in image-processing tasks \cite{Capece-2017}, and it retrieves the information lost through the convolutions of the layer groups. 
This information can be retrieved to help the decoder to reconstruct the output image. The proposed \fixme{encoder-decoder structured neural network} works in the RGB (red, green, blue) domains only, and it is able to recreate similar input images of faces, but in a different light mode.

Starting from the VGG-16 output tensor, to reconstruct the output image, we use deconvolutional layers \cite{Deconvolution}, which transpose the convolutional layers.

\subsection{Training}
Our CNN was trained to minimize the loss function using an algorithm called the Adam Optimizer \cite{kingma2014adam}. This algorithm is a stochastic gradient descent with momentum (SGDM) variant\cite{rumelhart1986learning}, which manages in a different way the problem of setting the learning rate. The choice of the learning rate can influence the CNN training convergence because a high value can lead to a possible divergence, while a very low value can lead to a slow convergence. SGDM addresses this problem by updating weights through a linear combination of the gradient and previous updates. Adam is an SGDM variant, which is based on two other well-known ones, called AdaGrad\cite{duchi2011adaptive} and RMSProp\cite{tieleman2012lecture}. These are classified in the category of adaptive algorithms because they adapt the learning rate for each of the parameters, leading to better convergence results. In particular, Adam combines the advantages of the methods mentioned above: It adapts the learning rate based on the first and second gradient moments. In our Adam configuration, the initial learning rate is set to $10^{-5}$, while $\beta_1$ and $\beta_2$, called the forgetting factors, are left at the default values, 0.9 and 0.999, respectively. A parameter, $\epsilon$, used to avoid divisions by zero, is set to $10^{-8}$\revision{, and finally the mini-batch size was set as 4 due to the input image dimension and the GPU capability}. 

To increase the generalization level and to compensate for the amount of the training data available, we initialized the weights of the VGG-16 encoder using a pre-trained model, which is used for face recognition \cite{Parkhi15}, exploiting the transfer learning concept\cite{Goodfellow-et-al-2016}. The model was trained using a very large-scale dataset that consists of 2.6 million faces belonging to 2600 identities (about 1000 images for each identity), using four GPU Titan Blacks. The input images of the pre-trained model have a resolution of $224\times 224$ pixels, from which was subtracted the mean of the training set's images. 

The weights of our decoder were initialized using a continuous probability distribution called a truncated normal distribution\cite{burkardt2014truncated}, which ensures that the weights initialization has mean zero and unit standard deviation. This avoids increasing or dissolution of the gradient, which otherwise leads to critical errors during training. The weights of the last decoder layer were initialized using Xavier Initialization \cite{glorot2010understanding}, which ensures that the weights are neither too large nor too small and that the signal passing through the neural network is propagated accurately. This prevents the signal from being amplified or reduced too much due to excessively large or small weight initialization. 


\subsection{Problem Encoding}\label{problem_encoding}
\label{sec:encoding}
\begin{figure}[!ht]
	\centering
	\includegraphics[width=1.0\textwidth]{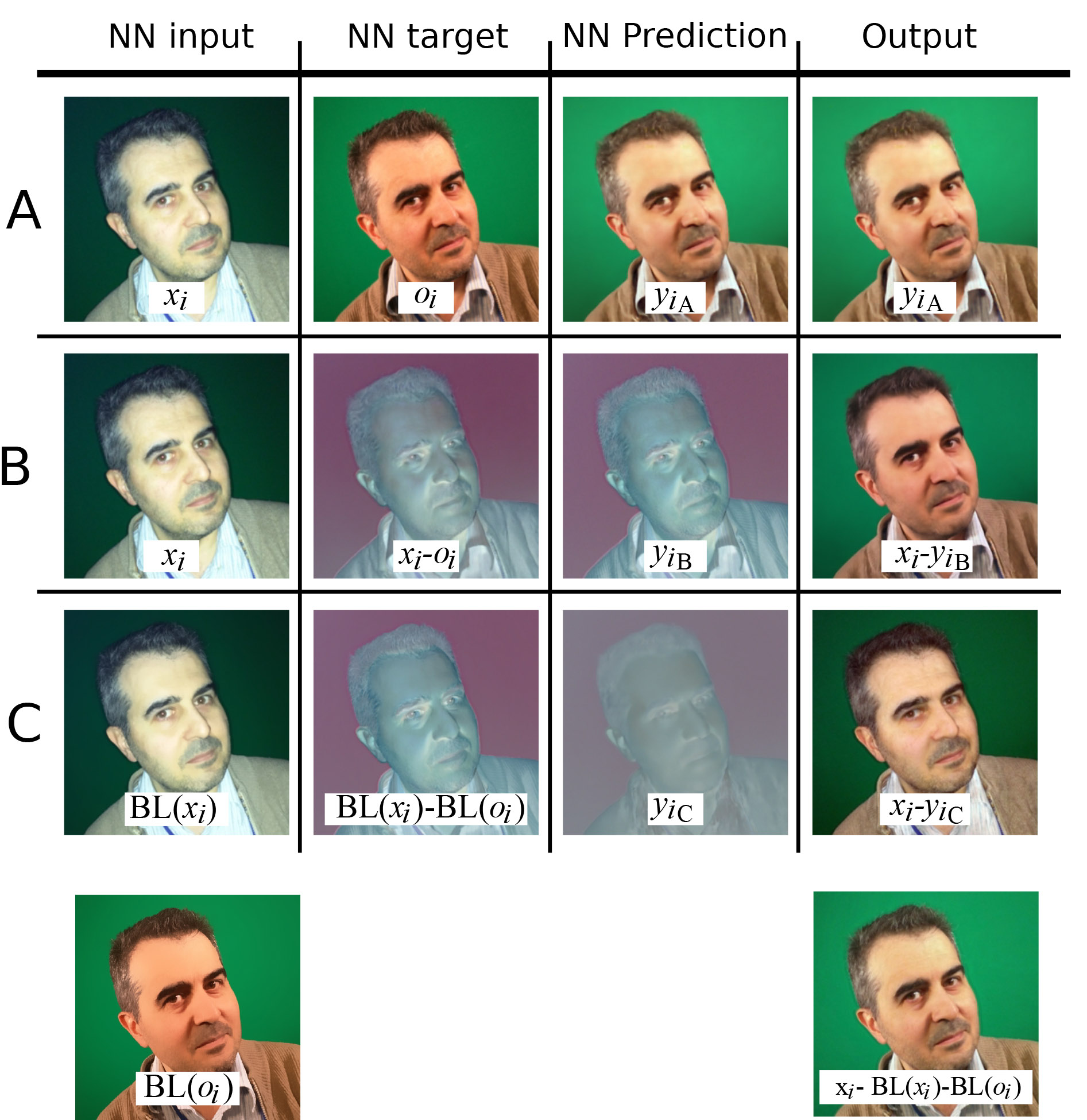}
	
	

\caption{\label{fig:encoding} Each row of the table shows a possible (and tested) encoding of the problem. For each row, the neural network (NN) input, target, prediction, and algorithm output are shown. The last row shows the bilateral filter applied to the uniform lit image (left) and the best achievable result with approach \textbf{C}.}
\end{figure}

The network can be used in more than one way to achieve our goal. 
The table in Figure~\ref{fig:encoding} illustrates three alternative encodings of the problem, showing the neural network input, target, prediction, and algorithm output for each one. We indicate with $x_i$ the flash image, $o_i$ the uniformly lit image, and $\mathbf{{y_i}_{[A|B|C]}}$ the network prediction.

The most straightforward solution, shown in the first row of Figure~\ref{fig:encoding}, consists of training the neural network by  providing $x_i$ as the input, $o_i$ as the target, and, once trained, to take the network prediction $\mathbf{{y_i}_{A}}$ as the final output.
With this setting, the network converged and gave good results in terms of colors and chromaticism. On the downside, the predictions were blurred, and small misalignments of facial expressions between $x_i$ and $o_i$ (e.g., eyes closed/open, the position of the lips, and other facial landmarks) created very visible artifacts in the predicted images. 

To reduce the blur in the images, we can train the network, giving as input the flash image and as the target the difference between $x_i$ and $o_i$. This way, the artifacts due to the alignment of facial expressions have been greatly reduced  and the training has been simplified. Through this approach, shown in the second row of Figure~\ref{fig:encoding}, the results visibly improved, but the  blur was not entirely removed.


Inspired by these results, we chose an encoding that decouples high-frequency details such as hairs of facial features from low-frequency characteristics such as the global skin tone.
We employed an accelerated version of the \emph{Bilateral Filter}\cite{Banterle+2012}, which is a nonlinear  filter that is ubiquitously used to smooth images preserving the edge features. Through this filter, we remove the input and the ground truth image's high frequencies, and we compute the distance between two images, normalizing it to $[0,1]$. The network's input is the filtered flash image, and the target is the distance  between the filtered input and the filtered ground truth:

\begin{equation}\label{eq1}
t_i = \frac{[\newfixme{BL}(x_i, \sigma_s, \sigma_r) - \newfixme{BL}(o_i, \sigma_s, \sigma_r)] + 1}{2} \quad x_i, o_i, t_i \in [0, 1]
\text{,}
\end{equation}

\noindent where \newfixme{$BL$} is the bilateral filter operator, $\sigma_s=16$ is the spatial sigma value, and $\sigma_r=0.4$ is the range sigma value. \revision{These parameters were selected from guidelines parameters proposed by Durand and Dorsey~\cite{Durand02} for separating low and high frequencies with a bilateral filter. Indeed, they suggested that $\sigma_s$ needs to be equal to the 2\% of the size of the maximum image dimension, and $\sigma_r = 0.4$. In our case, we slightly increased $\sigma_s$ to 3\% in order to increase quality when computing the filter with the used approximation~\cite{Banterle+2012} because samples are drawn proportionally to $\sigma_s$ in this approximation.}

The final reconstructed image is computed as:

\begin{equation}\label{eq1_}
pred_i = x_i - 2y_i + 1\\
\text{,}
\end{equation}

\noindent where $y_i$ is the network output.

\subsection{Loss Function}\label{loss_function}
As mentioned above, we trained our neural network with images filtered using the bilateral filter as the input, and the distance between input and ground truth filtered with the same filter as the target. The aim was to preserve the low frequencies and retrieve them in a subsequent step from the original non-filtered image. For this reason, we minimized the distance between the low frequencies of the input and ground truth. The objective function is therefore defined as follows:

\begin{equation}\label{eq2}
L(y_d,t_d) = \frac{1}{3N}\sum_{i}\biggl(({y_d}_i - \mathbb{E}[{y_d}_i]) - ({t_d}_i - \mathbb{E}[{t_d}_i])\biggr)^2 
\text{,}
\end{equation}
\noindent where
\begin{equation}\label{eq3}
\begin{split}
{y_d}_i = BL(x_i, \sigma_s, \sigma_r) - 2y_i + 1\\
{t_d}_i = BL(x_i, \sigma_s, \sigma_r) - 2t_i + 1
\end{split}
\text{.}
\end{equation}

In more specific terms, \newfixme{$N$ is the number of pixels}, $BL(x_i, \sigma_s, \sigma_r)$ is the CNN input, $x_i$ is the flash image, $y_i$ is the predicted difference of the CNN, and $t_i = BL(x_i, \sigma_s, \sigma_r) - BL(o_i, \sigma_s, \sigma_r)$, where $o_i$ is the ground truth. The bilateral filter's arguments are the same in equation (\ref{eq1}). Replacing equation (\ref{eq3}) in equation (\ref{eq2}), and by simplifying and exploiting the linear property of the mean, the objective function can be rewritten as:

\begin{equation}\label{eq4}
L(y,t) = \frac{4}{3N}\sum_{i}\biggl((t_i - y_i) + \mathbb{E}[y_i - t_i]\biggr)^2 
\text{.}
\end{equation}

To avoid negative values affecting the CNN convergence due to activation functions, we normalized the target difference image in the range [0\ldots1] (\ref{eq1_}). In particular, since \textit{ReLU} and \textit{LeakyReLU} are non-saturating nonlinear activation functions\cite{AlexNet} \cite{sun2015deeply}, they tend to eliminate completely or partially the negative output values of each layer, leading to faster convergence than saturating nonlinear activation functions such as $\tanh$, which lead to longer training times and a slower convergence. Furthermore, it has been shown that rectified units are much more efficient for tasks concerning images\cite{Nair_Vinod} \cite{glorot2011deep}. Therefore, the network will perform predictions in the $[0, 1]$ range; for this reason, the values are reported in the $[-1, 1]$ range and subsequently subtracted from the input values. If we consider the bilateral filter function and perform a further substitution, we can insert the original non-filtered images into equation (\ref{eq3}), and the objective function can be explicitly rewritten as:

\begin{equation}
\begin{split}
L(y,x,o) = \frac{4}{3N}\sum_{i}\biggl((BL(x_i, \sigma_s, \sigma_r) - BL(o_i, \sigma_s, \sigma_r) - y_i) +\\
 + \mathbb{E}[y_i - BL(x_i, \sigma_s, \sigma_r) + BL(o_i, \sigma_s, \sigma_r)]\biggr)^2
\end{split}
\end{equation} 

Mean subtraction is performed for each channel of each image pixel by pixel only in the evaluation phase of the objective function, to centralize the data and to distribute the weights of each image across the training in a balanced manner, so that each image gives the same contribution to the training and does not have more or less important than the others. This normalization operation is performed differently for every single image, compared with the classic method of centralizing the data, which involves subtracting from each image the mean computed across the whole training dataset. Because our problem is confined to a specific domain, in which the data are stationary and the image lighting parameters are well defined and always the same both for the input and for the output, we subtracted the mean for each single image, which was computed on the same image to remove the average brightness or intensity from each pixel.

\section{Experimental Setup -- Dataset Creation }\label{dataset_creation}
We performed neural network training using pairs of photos taken with a 13 Megapixel Nexus 6 smartphone camera.
The photographs were taken in a studio equipped with four Lupoled 560 lamps to provide uniform illumination.
\revision{We choose not to use the synthetic data for pretraining because the variety and quality of the lighting artifacts that we aimed to capture and correct needed a high number of high quality photorealistic 3D models. The generation of such models was considered less practical and safe that direct acquisition of real examples.} For each pose, a photograph was taken with the lamps on, which were then immediately switched off, and a second photograph was taken with the smartphone flash only. Because the switching off the lamp imposes a significant delay between the two shots (about 400ms with our lamps), the pose of the face between the first and  the second may change significantly. 

To reduce the problem of face misalignment, we performed an \textit{affine} alignment (i.e., translation, rotation, scale, and shear) using the MATLAB Image Processing Toolbox\texttrademark. In particular, we considered the non-flash image as the misaligned image ($M$) and the flash image as our reference ($R$).
Because the images have different lighting conditions, we employed a \textit{multimodal} metric\cite{ raghunathan2005image} \cite{mattes2001nonrigid} and optimizer\cite{ 845174}. Once the geometric transformation was estimated, we applied it to $M$, obtaining $M^\prime$, which is a better alignment to $R$.
Note that a limitation of this approach is that misalignments remain between open and closed eyes.
After affine registration, we identified the face of a subject in $M$, $M^\prime$, and $R$ using the Face Recognition API \footnote{\url{https://github.com/ageitgey/face_recognition}}, which returns a bounding box for the photograph. Each bounding box is used to crop the image, and then the image is downsampled to $512 \times 512$. \revision{As the aim of CNN is to automatically extract the input image features, preserving globally the spatial characteristics of the images such as the edges is very important especially in the first layers of the CNN~\cite{Goodfellow-et-al-2016}. In this way, the spatial structure of the images are preserved and equivariance to the translation of CNN is ensured. 
}


We collected about $495$ pair of photos of $101$ of people (both females and males) in different poses.
Then, we augmented the dataset in three ways. First, through $5$ rotations from $-20$ to $20$ degrees around the center of face bounding box, using a $10$ degree step. Second, by cropping the image to the face bounding box and rescaling to original image size. Finally, images are flipped horizontally.
Altogether, we augmented the initial set of examples by a factor $20$, so our training is performed with $9.900$ images with a $3120 \times 4160$ resolution ($13$ Megapixel).

\section{Results}
\label{sec:results}
We evaluated the results in validation and test sets~\revision{(see Figure~\ref{fig:teaser})}. In particular, the CNN was trained using pairs of images with a resolution of $512 \times 512$ in about five days using a NVIDIA Titan Xp GPU and by performing 62 epochs and about 458,000 backpropagation iterations. We interrupted the training when the value of the loss function computed on 1,500 images reached a low level of approximately 0.0042. To evaluate the similarity accuracy between the current prediction and ground truth, we computed the percentage difference between the two images as follows:

\begin{equation}\label{eq6}
acc = 100 - \bigg(\frac{100}{3 w(I) h(I)}\sum_{i}\sum_{c}\mid I_c - \tilde{I}_c \mid \bigg)
\end{equation}

\begin{figure}[!ht]
	\centering
	\includegraphics[width=0.85\columnwidth]{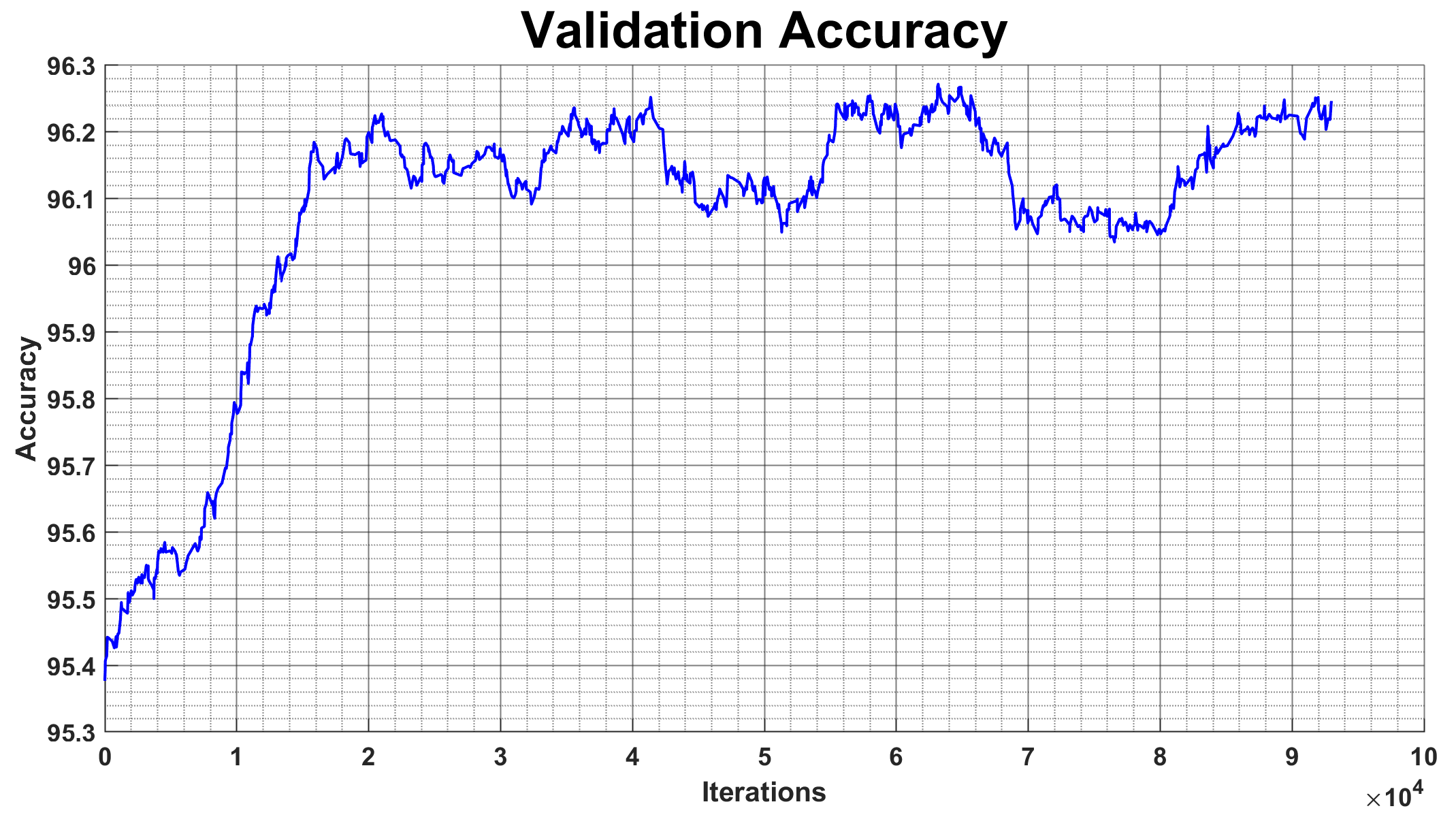}
	\caption{Accuracy trend in the validation set at the end of training the CNN after 62 epochs and 458,000 iterations performed in 5 days.}
\label{fig:accuracy}
\end{figure}

\noindent where $I = t_d, \tilde{I} = y_d, w(I) = width(I)$, and $h(I) = height(I)$. After the training step, we obtained an accuracy value of 96.2\% (Figure \ref{fig:accuracy}). In the test phase, we evaluated our approach using 740 test images, obtaining a loss-test value of 0.0045 and an accuracy value of 96.5\% (Table \ref{tab1}).

\subsection{Comparison with Reconstructed Ground Truth}
\label{sec:gt}
As explained in Section~\ref{sec:encoding}, the result provided by our pipeline
is obtained by subtracting the CNN prediction from the original input image, allowing us to retrieve the high frequencies, which had been lost due to the bilateral filter. It follows that even for a loss function $L(y,x,o) = 0$ the exact ground truth can never be reconstructed. Furthermore, and more importantly, the  misalignements due to pose changes between the flash and non flash photos would dominate when computing the input and output image differences. For these reasons, we introduce a preconditioning operator on the ground truth as:

\begin{equation}\label{eq7}
\overline{o}_i = x_i - 2t_i + 1
\end{equation}

\begin{figure*}[!ht]  
	\centering
    \setkeys{Gin}{draft=false}
	\includegraphics[width=\textwidth]{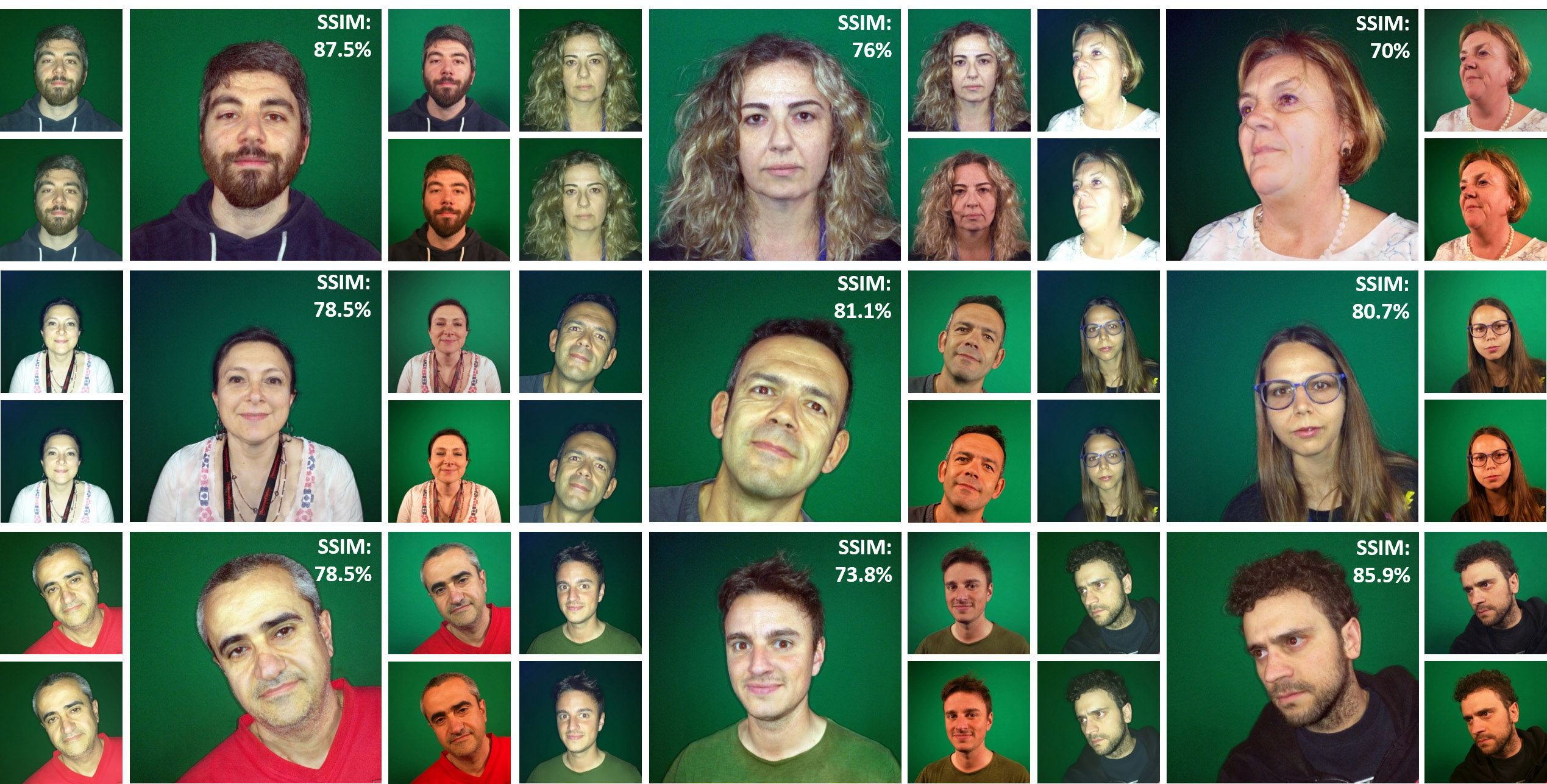}
	\caption{\label{fig:results}Samples of validation data. Each group of images is composed of: the original image taken with the smartphone flash (top left); the flash image to which the bilateral filter was applied (bottom left); the image reconstructed by the difference prediction of the CNN (center); the ground truth reconstructed (top right); the ground truth (bottom right). \revision{The SSIM was computed by comparing the central image of each group and the image at top right.}}

\end{figure*}

\begin{figure}[!ht]
 \setlength{\tabcolsep}{1pt}
\begin{tabular}{ccc}
Input & Output & Ground Truth\\
\includegraphics[width=0.31\textwidth]{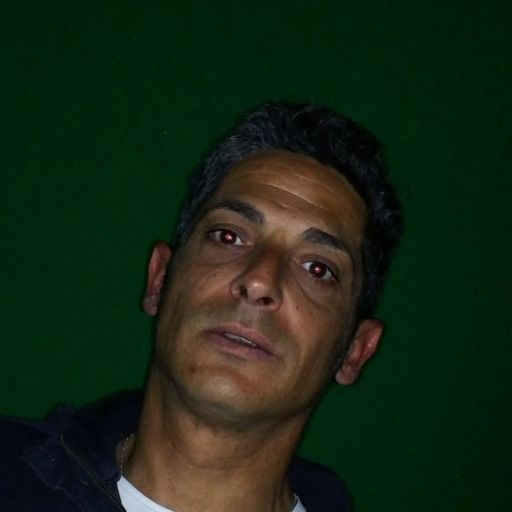}\label{fig:1_input}&
\includegraphics[width=0.31\textwidth]{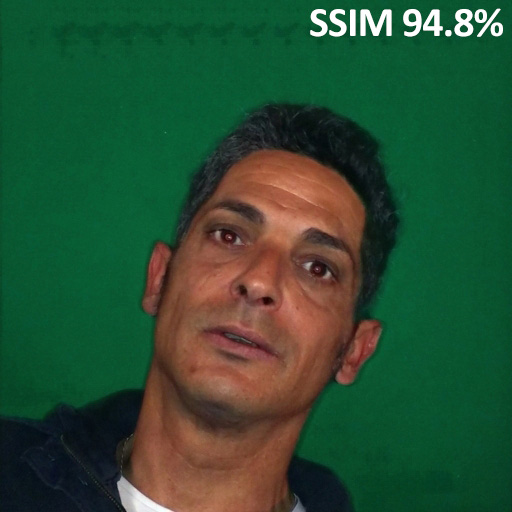}\label{fig:1_prediction}&
\includegraphics[width=0.31\textwidth]{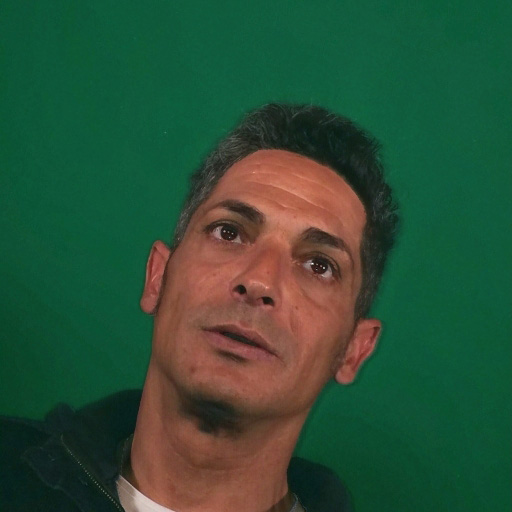}\label{fig:1_gt}\\

\includegraphics[width=0.31\textwidth]{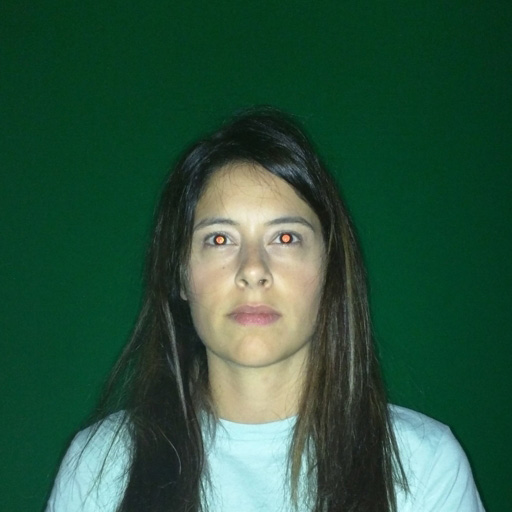}\label{fig:2_input}&
\includegraphics[width=0.31\textwidth]{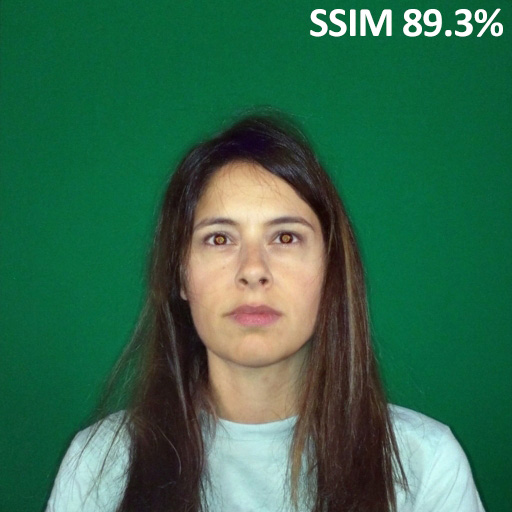}\label{fig:2_prediction}&
\includegraphics[width=0.31\textwidth]{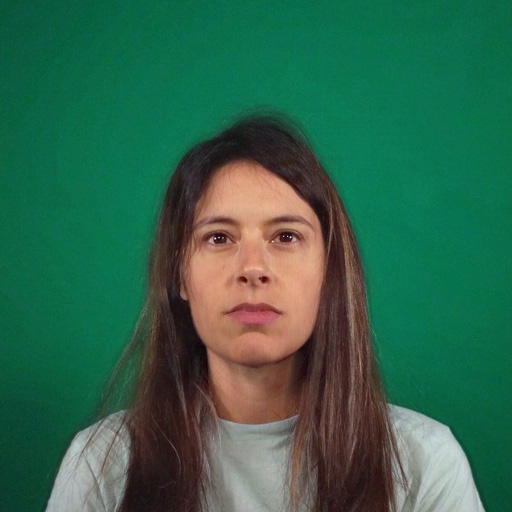}\label{fig:2_gt}\\

\includegraphics[width=0.31\textwidth]{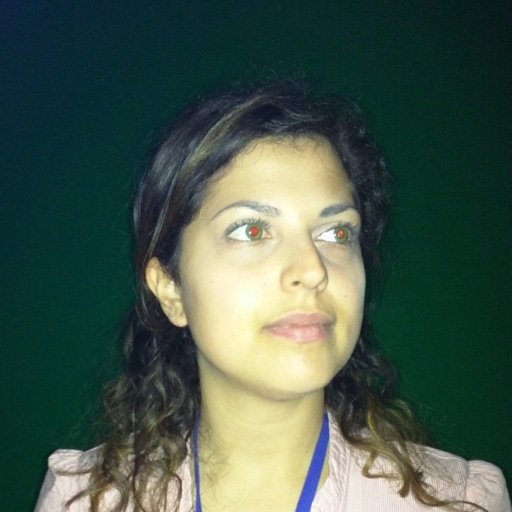}\label{fig:3_input}&
\includegraphics[width=0.31\textwidth]{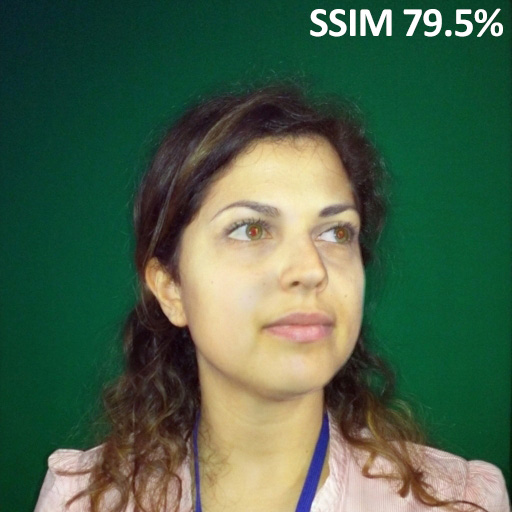}\label{fig:3_prediction}&
\includegraphics[width=0.31\textwidth]{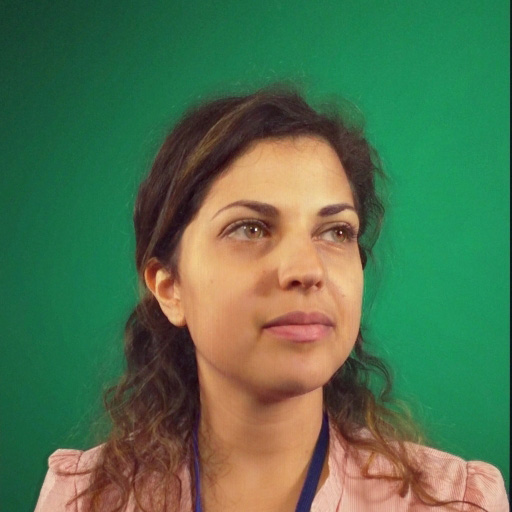}\label{fig:3_gt}\\
\end{tabular}	

\caption{\label{fig:training}Training set example. For each row: the original input (left); the image reconstructed by CNN prediction (center); the ground truth reconstructed (right). \revision{The SSIM was computed by comparing the center images and the right images.}}
\end{figure}

\begin{figure}[!ht]
 \setlength{\tabcolsep}{1pt}
\begin{tabular}{ccc}
Input & Output & Ground Truth\\

\includegraphics[width=0.31\textwidth]{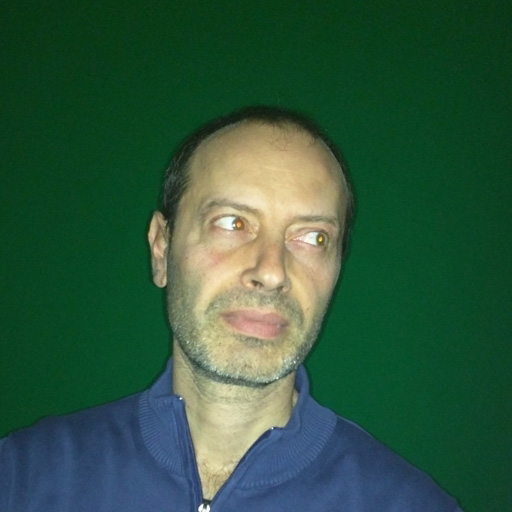}\label{fig:1_t_input}&

	\includegraphics[width=0.31\textwidth]{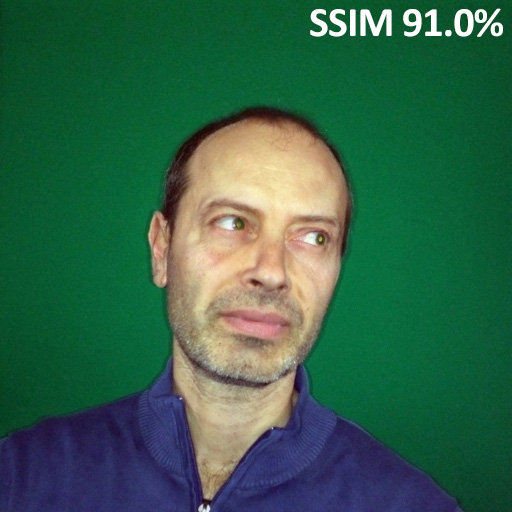}\label{fig:1_t_prediction}&
	
	\includegraphics[width=0.31\textwidth]{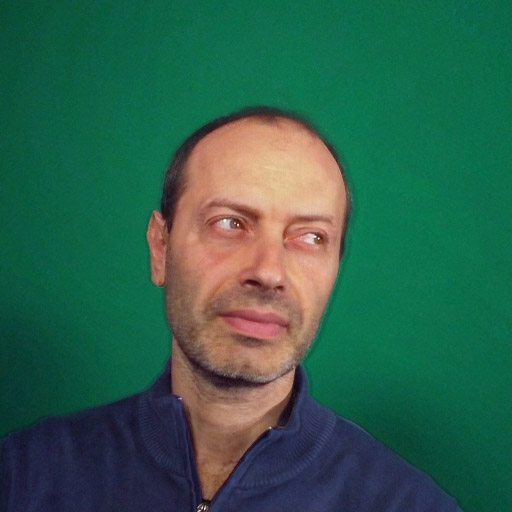}\label{fig:1_t_gt}\\

	\includegraphics[width=0.31\textwidth]{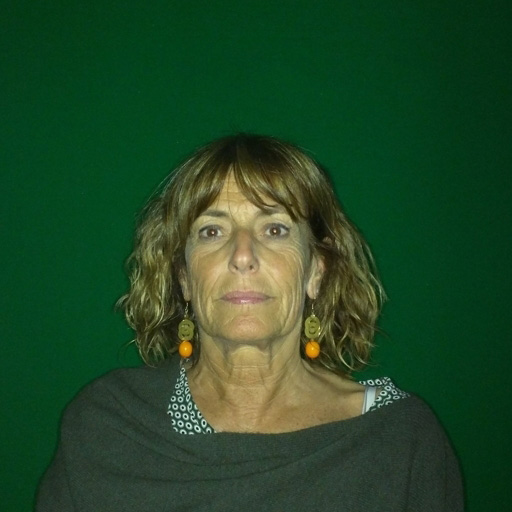}\label{fig:2_t_input}&

	\includegraphics[width=0.31\textwidth]{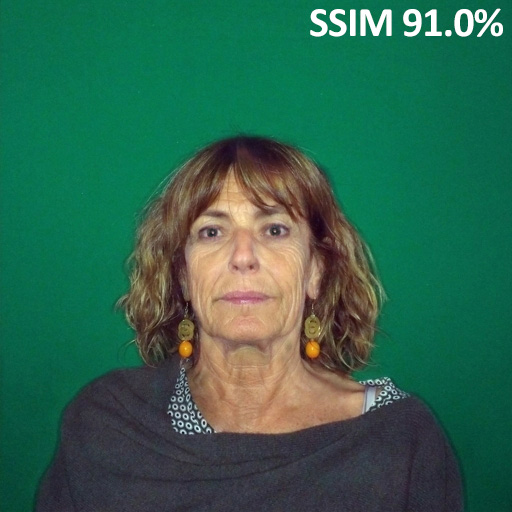}\label{fig:2_t_prediction}&
	
	\includegraphics[width=0.31\textwidth]{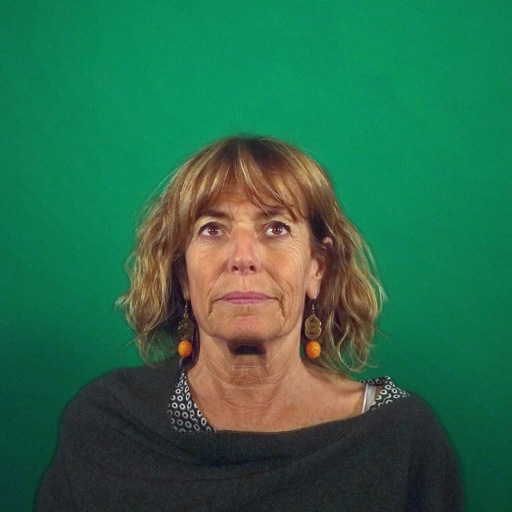}\label{fig:2_gt_}\\

	\includegraphics[width=0.31\textwidth]{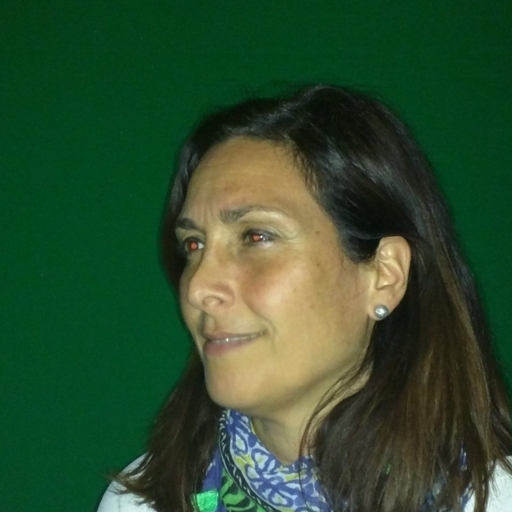}\label{fig:3_t_input}&

	\includegraphics[width=0.31\textwidth]{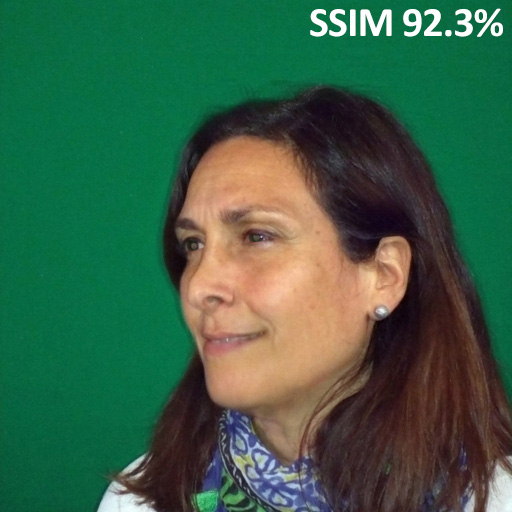}\label{fig:3_t_prediction}&
	
	\includegraphics[width=0.31\textwidth]{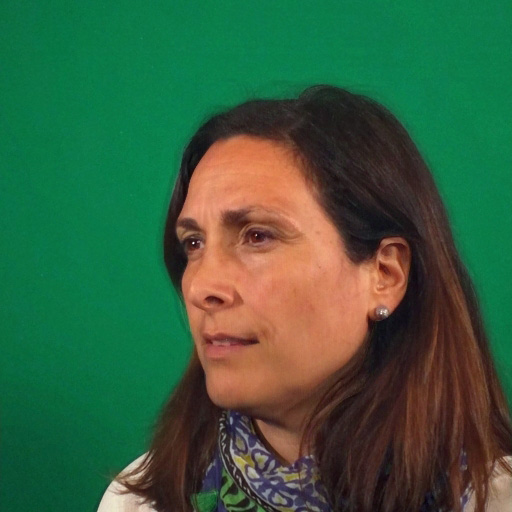}\label{fig:3_t_gt}\\
\end{tabular}	

\caption{\label{fig:test} Test set example. For each row: the original input (left); the image reconstructed by CNN prediction (center); the ground truth reconstructed (right). \revision{The SSIM was computed by comparing the center images and the right images.} \revision{The SSIM was computed on the images without removing the red-eye artifact.}}
\end{figure}

\begin{figure}[!ht]
 \setlength{\tabcolsep}{1pt}
\begin{tabular}{cc}
Flash Image & Result\\
    \includegraphics[width=0.495\textwidth]{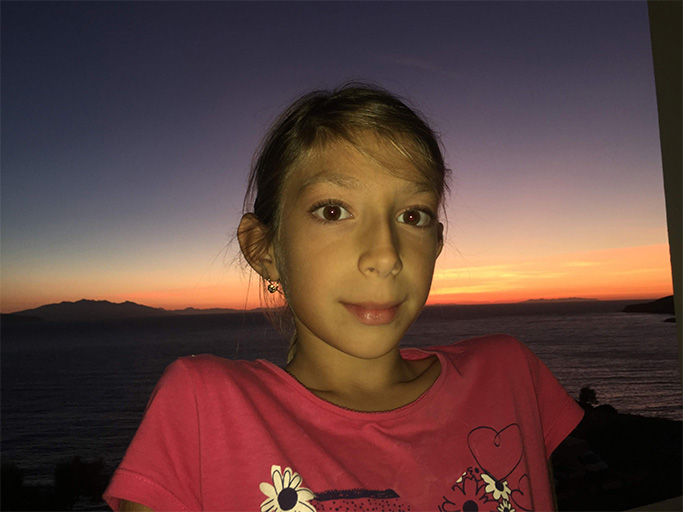}\label{fig:real_input_1}&
	\includegraphics[width=0.495\textwidth]{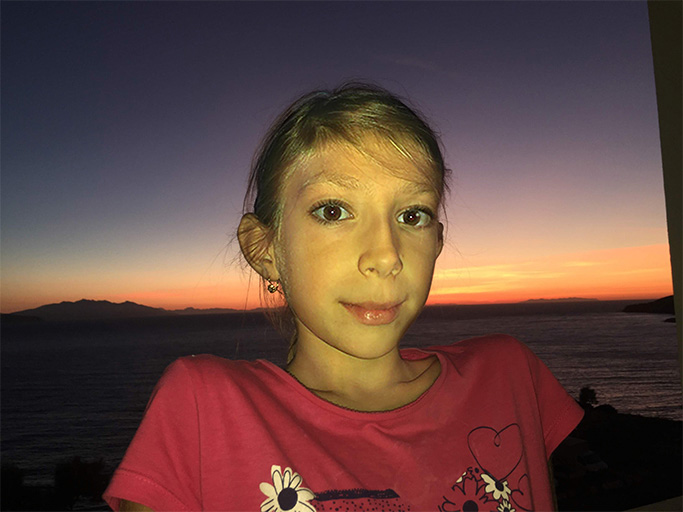}\label{fig:real_pred_1}\\
	\includegraphics[width=0.495\textwidth]{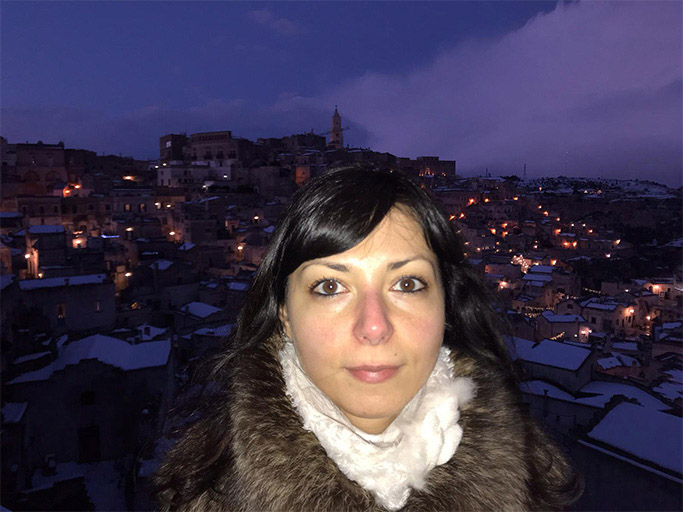}\label{fig:real_input}&
	\includegraphics[width=0.495\textwidth]{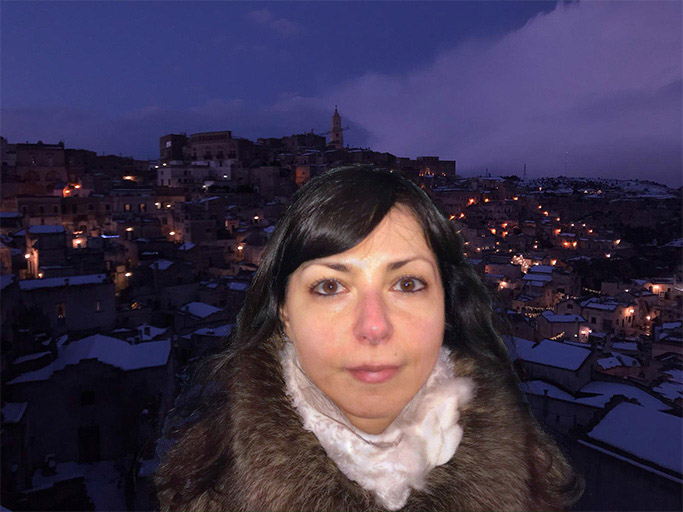}\label{fig:real_pred}
\end{tabular}	

\caption{\label{fig:real_result}Two example of a real images. The subjects were detected using Photoshop Subject Detection tool and were forwarded to our neural network with a green background. The results were blended with the original images.  }
\end{figure}

\revision{where $t_i$} is the target difference, as explained in Section \ref{loss_function}. Figure \ref{fig:results} shows the final result for some examples of the validation data, while Figures \ref{fig:training} and \ref{fig:test} show, similarly, the reconstruction of example images belonging to the training and test sets, respectively. In particular, it can be seen in Figures \ref{fig:results} and \ref{fig:test} that the information lost because of the flash, such as hairs, beard and skin color, are retrieved through the CNN's prediction. The shadow cast by the subject is also corrected by the CNN to reduce the highlights due to the camera flash. By reconstructing the final image through the non-filtered input, we retrieved the high frequencies resolving the blur problem. We  evaluated the data using the Structural Similarity Index (SSIM)\cite{SSIM} for each subset of the dataset. In particular, as can be seen from \revision{Figures~\ref{fig:results},~\ref{fig:training}, and~\ref{fig:test}, which show comparisons between ground truth and the reconstructed predictions,} 
the SSIM average value is around 80\% for the validation set, around 92\% for the test set, and around 94.5\% for the training set.
As a final step, we run a Red Eye Removal filter with GIMP to present the final results \revision{only on the test set images for aesthetic reasons; see Figure~\ref{fig:test}}. Please note that all metrics were run on images without removing the red-eye artifact \revision{(see Figure~\ref{fig:real_result})}.

\begin{table}[!ht]
  \centering
  \begin{tabular}{|p{2.5cm}|p{2.5cm}|p{2.5cm}|}
    \hline
     & Loss         & Accuracy \\ \hline
    Validation               & 0.0042 & 96.2\%   \\ \hline
    Test                     & 0.0045 & 96.5\%   \\ \hline
  \end{tabular}
  \newline\newline
  \caption{Loss validation and loss test after 62 epochs. This table also shows the maximum accuracy achieved in both phases.}\label{tab1}
\end{table}

 
	 
		

\newpage
\subsection{Comparison with HDRNet}\label{HDRNET:COMP}
We compared our approach to HDRNet by Gharbi et. al\cite{gharbi2017deep}. This work is based on the use of a CNN inspired to the bilateral grid processing \cite{Chen:2007:REI:1276377.1276506} and to local affine color transforms.
HDRNet is designed to learn any image operator and hence is a suitable candidate to remove flash artifacts from photographs. 
In a first experiment, we trained HDRNet end-to-end with our dataset, by feeding the network with input and target images. We used the same parameters, times and training algorithm proposed by the authors and trained the  network for 48 hours. We obtained a stable loss value of 0.0031. 

\begin{figure}[!ht]
 \setlength{\tabcolsep}{1pt}
\begin{tabular}{cccc}
	Input & HDRNet & Our approach & Ground Truth\\
	{\includegraphics[width=0.245\textwidth]{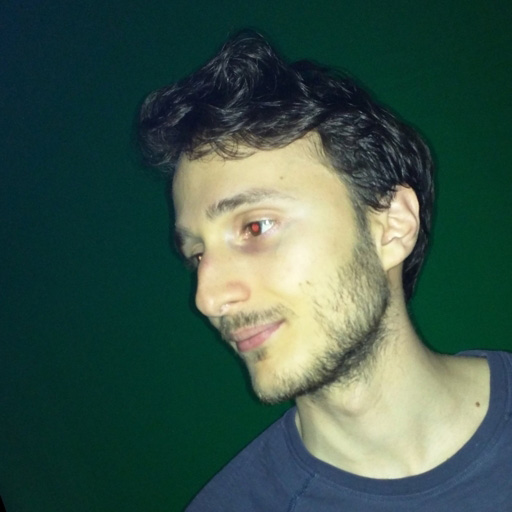}\label{fig:train_e-ep_input}}&
	{\includegraphics[width=0.245\textwidth]{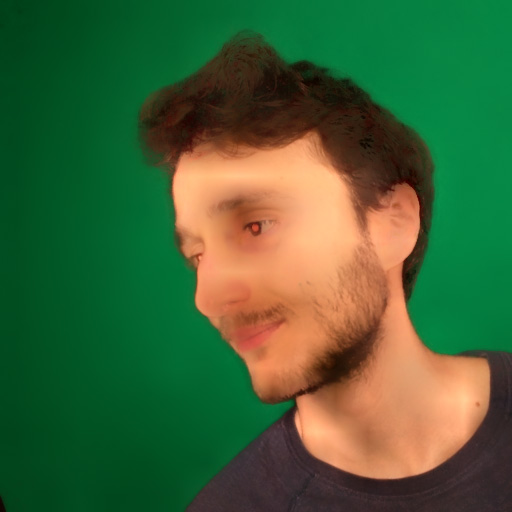}\label{fig:train_e-e_h}}&
	{\includegraphics[width=0.245\textwidth]{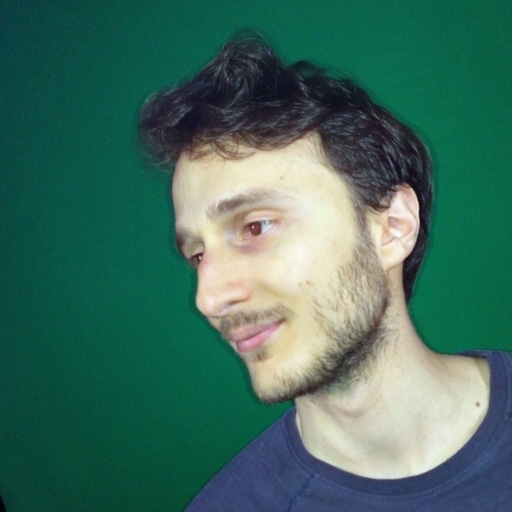}\label{fig:train_e-e}}&
	{\includegraphics[width=0.245\textwidth]{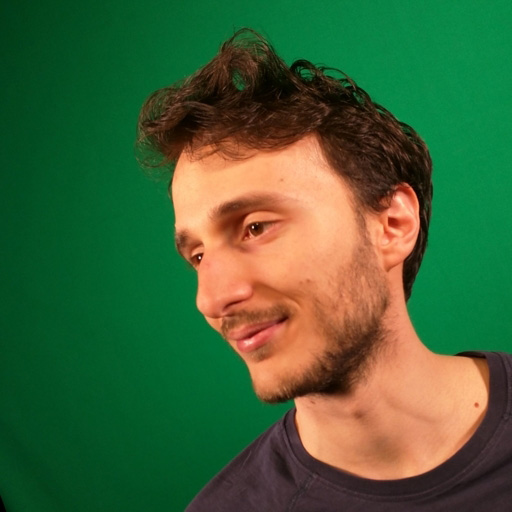}\label{fig:train_e-e_target}}\\
\end{tabular}	
\caption{\label{fig:comp_ee} A comparison between HDRNet end-to-end training and our approach. }
\end{figure}

Figure~\ref{fig:comp_ee} shows the comparison with our approach. Although HDRNet is capable of approximating the colors of the ground truth image, the flash highlights remain substantially unchanged and the blurring is extremely high making the face unrecognizable. Note that this is also a consequence of the input and output image misalignments for which we introduced the preconditioning operator explained at the beginning of Section~\ref{sec:gt}. On the contrary, our result more closely matches the studio portrait preserving the high frequencies of the images such as hair and face traits. 

\begin{figure}[!ht]
\centering
	 \setlength{\tabcolsep}{1pt}
	\begin{tabular}{ccccc}
	&Input & HDRNet & Our approach & Ground Truth\\
	&  &   our encoding &   &  \\
	\rotatebox{90}{\ \ \ \ \ Training}&{\includegraphics[width=0.235\textwidth]{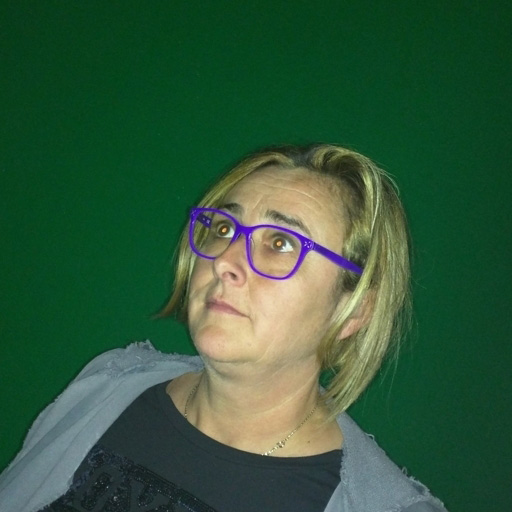}\label{fig:train_comp_input}}&
	{\includegraphics[width=0.235\textwidth]{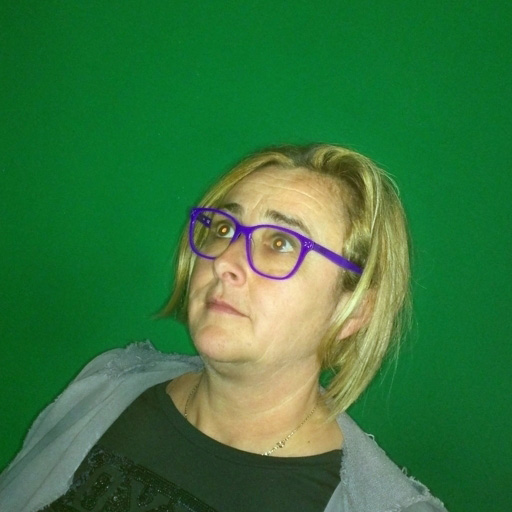}\label{fig:train_comp_h}}&
	{\includegraphics[width=0.235\textwidth]{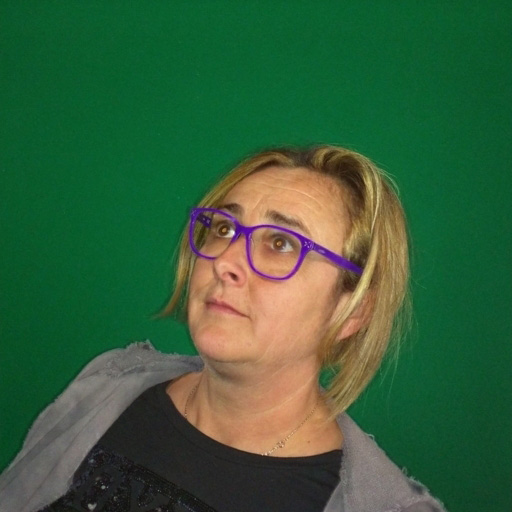}\label{fig:train_comp}}&
	{\includegraphics[width=0.235\textwidth]{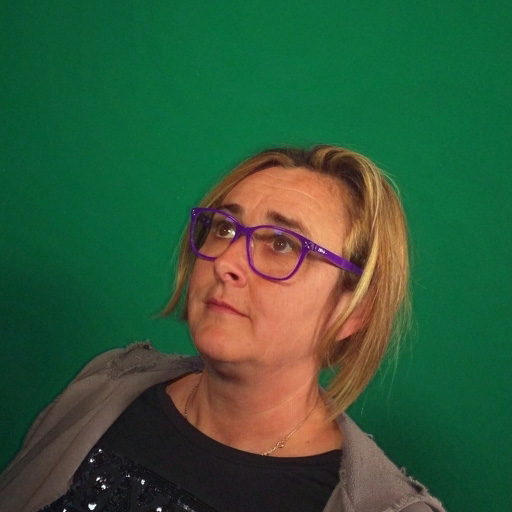}\label{fig:train_target}}\\
	
	\rotatebox{90}{\ \ \ \ Validation}&{\includegraphics[width=0.235\textwidth]{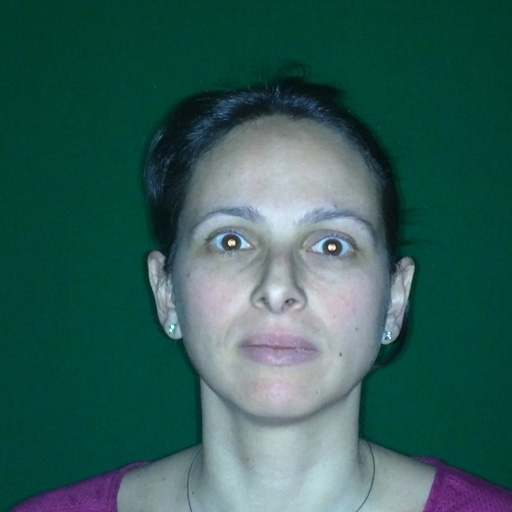}\label{fig:val_comp_input}}&
	{\includegraphics[width=0.235\textwidth]{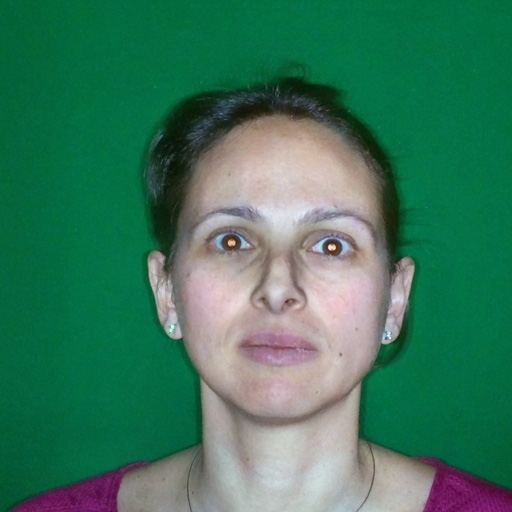}\label{fig:val_comp_h}}&
	{\includegraphics[width=0.235\textwidth]{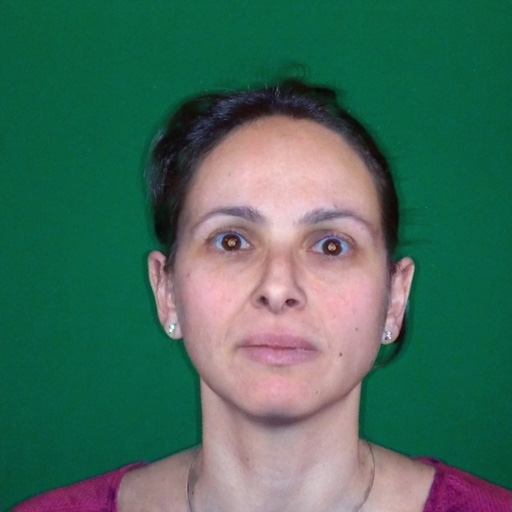}\label{fig:val_comp}}&
	{\includegraphics[width=0.235\textwidth]{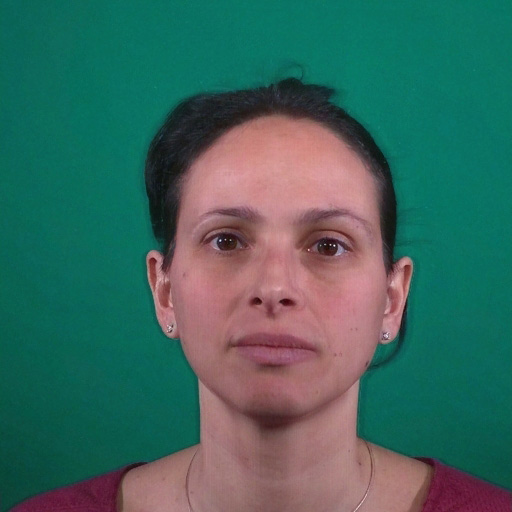}\label{fig:val_target}}\\
	
	\rotatebox{90}{\ \ \ \ \ \ \ Test}&{\includegraphics[width=0.235\textwidth]{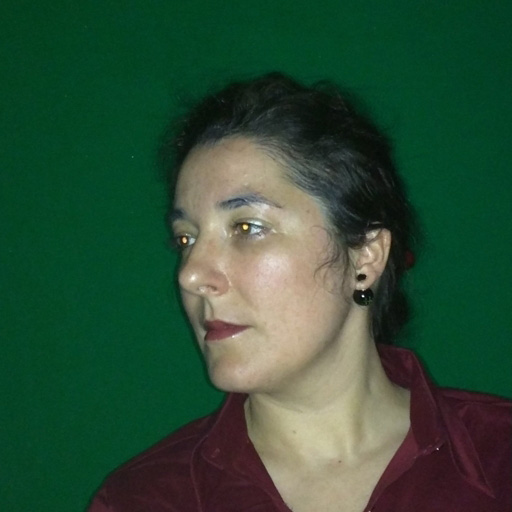}\label{fig:test_comp_input}}&
	{\includegraphics[width=0.235\textwidth]{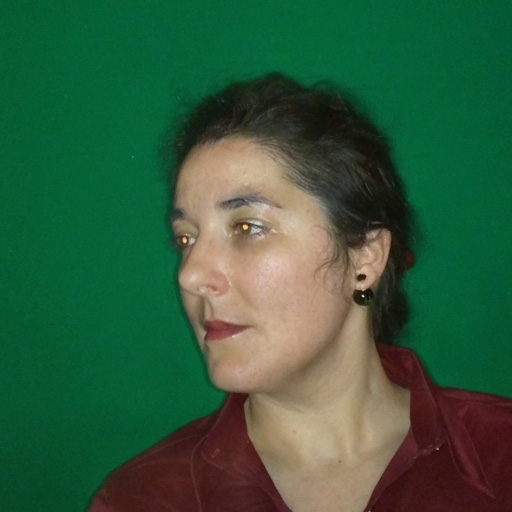}\label{fig:test_comp_h}}&
	{\includegraphics[width=0.235\textwidth]{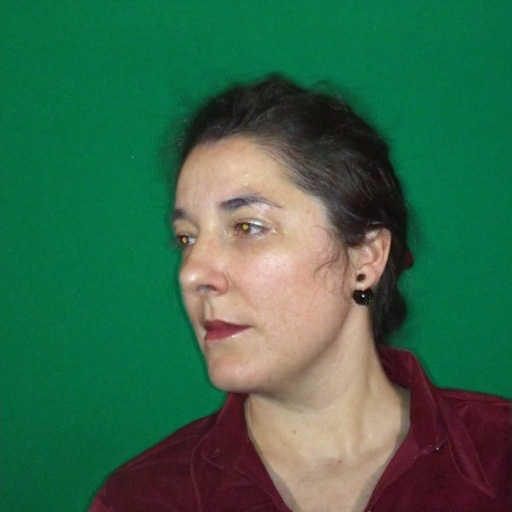}\label{fig:test_comp}}&
	{\includegraphics[width=0.235\textwidth]{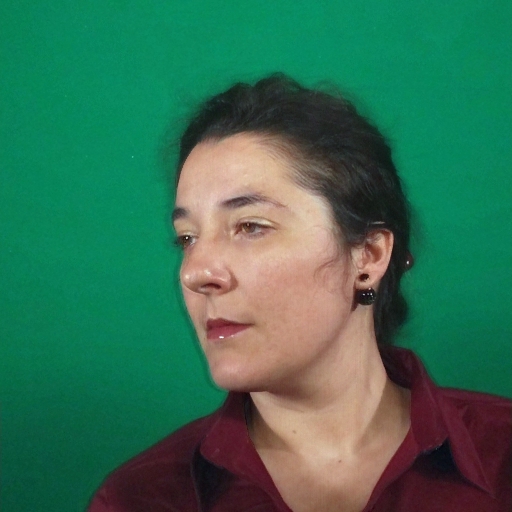}\label{fig:test_target}}\\
	\end{tabular}
	
\caption{\label{fig:comp} An example of comparisons between HDRNet (combined with our problem encoding) and our approach. We show one example from the training set, the validation set and the test set, respectively.}
\end{figure}

In a second experiment, we combined HDRNet with our encoding, that is, 
training HDRNet with the filtered images as input and the difference between the input and the filtered ground truth as target (see Section~\ref{problem_encoding}).
Even in this experiment, we used the parameters proposed by the authors and trained the network for over 48 hours, obtaining a stable loss value of 0.0007.
A few result samples of this experiment are shown in Figure~\ref{fig:comp}. From this figure, we can notice that HDRNet performance has dramatically improved by using our encoding. However, our full approach (that is, our encoding on our network) does a better job at removing the flash artifacts (i.e., highlight and shadows).

We compared the  two methods by using SSIM and the Peak Signal to Noise Ratio (PSNR) between the prediction of networks and the target images obtained a random sample of 30 images from each sub-set of the dataset (training set, validation set, and test set). The results are reported in the Table~\ref{tab_comp}. From this table, we can see that our approach results in  higher SSIM and PSNR values for all subsets.

\revision{We use the SSIM\cite{SSIM} index to compare the images because such metric can measure the similarity between two images in a way that is consistent with human eye perception. 
In addition, we use also the PSNR, because it is an approach that estimates the absolute error between two images. Furthermore, such metric is often used to compare the quality of reconstructed images, as in this case; i.e., the CNN output. }

Summarizing, we can claim both that our approach outperforms HDRNet w.r.t. the specific image operator that removes the flash artifacts, and that HDRNet benefits from our encoding strategy by producing better results than its native end-to-end configuration.


\begin{table}[!ht]
  \centering
  \begin{tabular}{|p{2.0cm}|p{2.0cm}|p{2.0cm}|p{2.0cm}|p{2.0cm}|}
    \hline
     & Our SSIM & HDRNet SSIM & Our PSNR & HDRNet PNSR \\ \hline
     
    Training                 & \textbf{91.55\%} & 73.93\% & \textbf{24.05 dB} & 19.71 dB   \\ \hline
    Validation               & \textbf{87.76\%} & 78.51\% & \textbf{20.42 dB} & 18.84 dB   \\ \hline
    Test                   & \textbf{89.80\%} & 82.92\% & \textbf{21.28 dB} & 19.02 dB   \\ \hline
  \end{tabular}
  \newline\newline
  \caption{Comparisons on samples of the training, validation, and test sets. In particular, the SSIM and PSNR values are computed on samples of 30 images extracted randomly from each dataset.}\label{tab_comp}
\end{table}
\newpage
\subsection{Comparison with Pix2Pix}
\begin{figure}[!ht]
 \setlength{\tabcolsep}{1pt}
\begin{tabular}{cccc}
	Input & Pix2Pix & Our approach & Ground Truth\\
	{\includegraphics[width=0.245\textwidth]{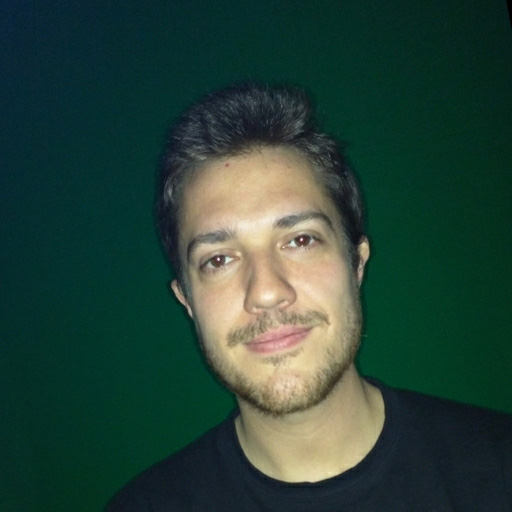}\label{fig:train_p2p_input}}&
	{\includegraphics[width=0.245\textwidth]{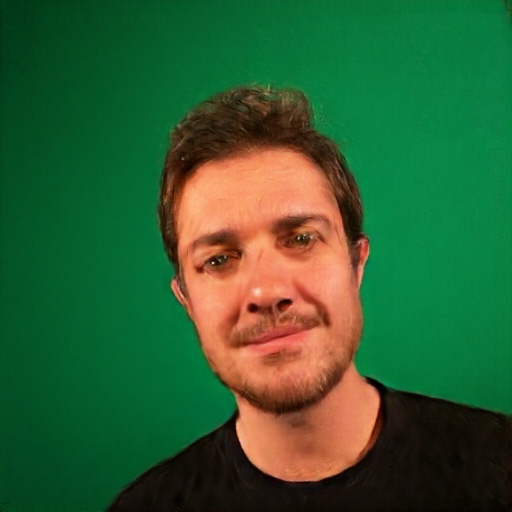}\label{fig:train_p2p_output}}&
	{\includegraphics[width=0.245\textwidth]{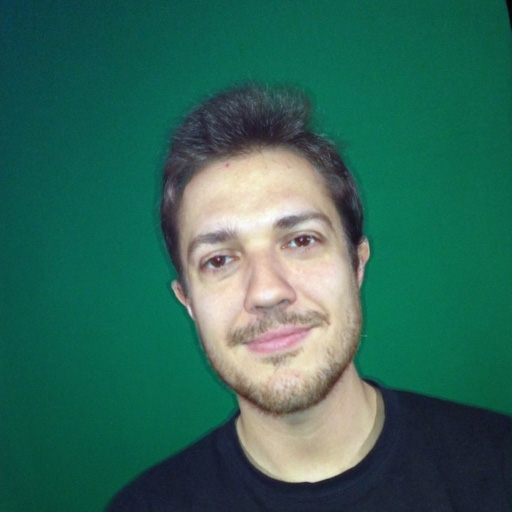}\label{fig:train_our_output}}&
	{\includegraphics[width=0.245\textwidth]{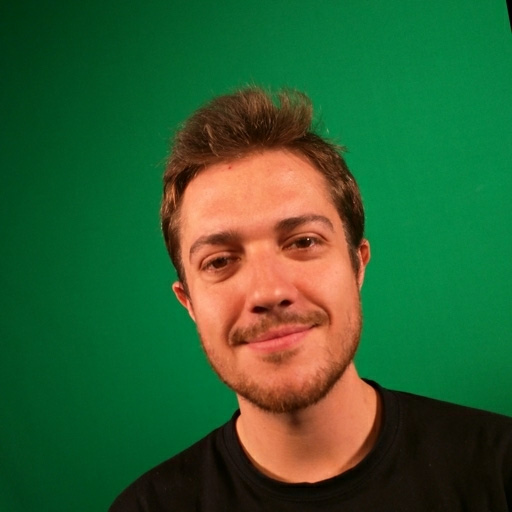}\label{fig:train_p2p_target}}\\
\end{tabular}	
\caption{\label{fig:comp_ee_p2p} \newfixme{A comparison between Pix2pix end-to-end training and our approach.}}
\end{figure}

\begin{figure}[!ht]
\centering
	 \setlength{\tabcolsep}{1pt}
	\begin{tabular}{ccccc}
	&Input & Pix2Pix & Our approach & Ground Truth\\
	&  &   our encoding &   &  \\
	\rotatebox{90}{\ \ \ \ \ Training}&{\includegraphics[width=0.235\textwidth]{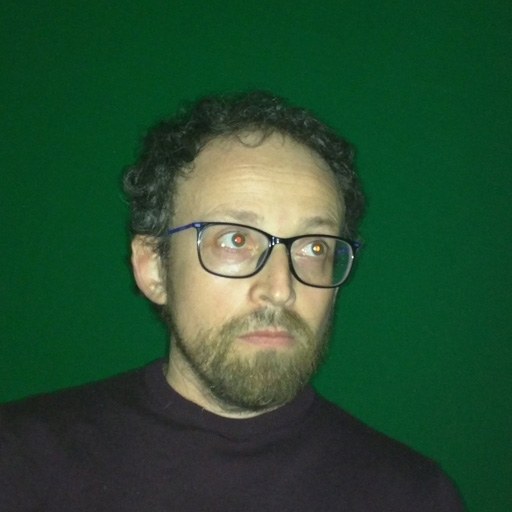}\label{fig:train_p2p_input_}}&
	{\includegraphics[width=0.235\textwidth]{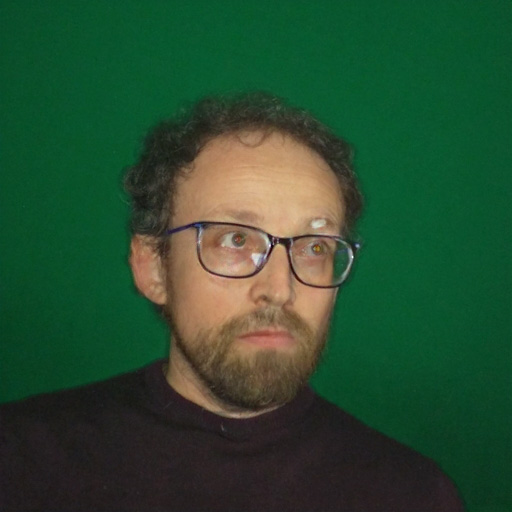}\label{fig:train_p2p_h}}&
	{\includegraphics[width=0.235\textwidth]{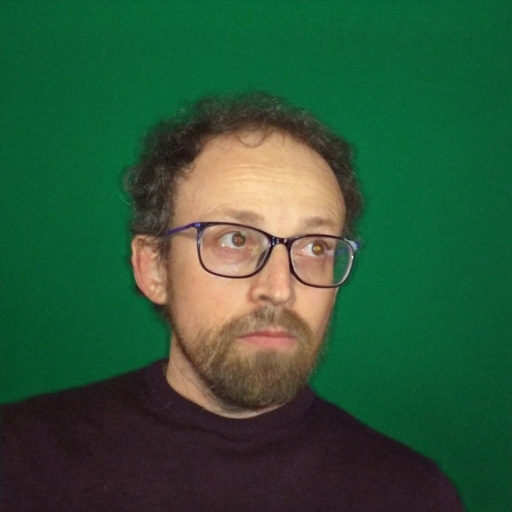}\label{fig:train_p2p}}&
	{\includegraphics[width=0.235\textwidth]{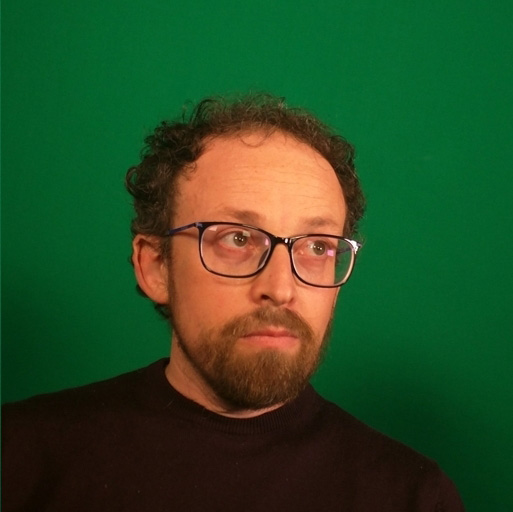}\label{fig:train_p2p_target_}}\\
	
	\rotatebox{90}{\ \ \ \ Validation}&{\includegraphics[width=0.235\textwidth]{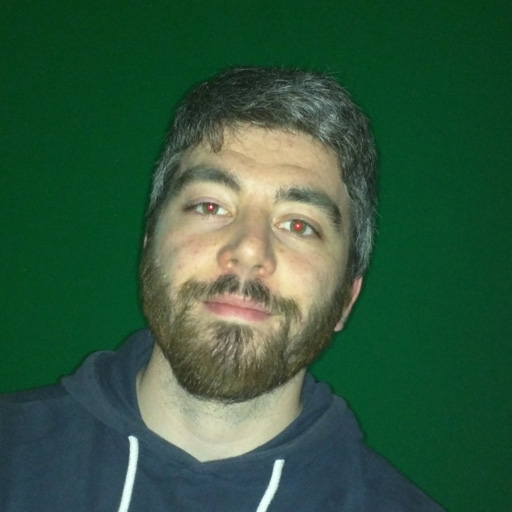}\label{fig:val_p2p_input}}&
	{\includegraphics[width=0.235\textwidth]{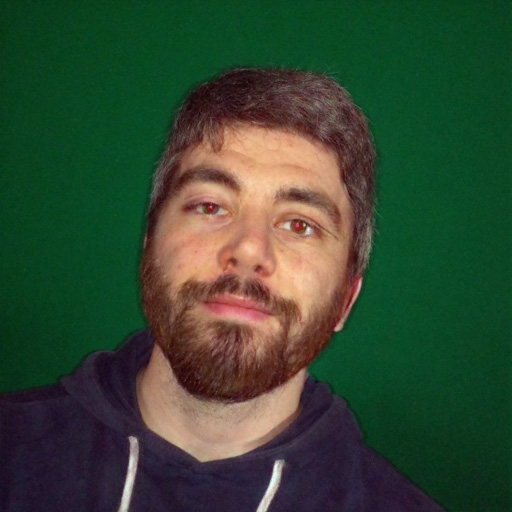}\label{fig:val_p2p_h}}&
	{\includegraphics[width=0.235\textwidth]{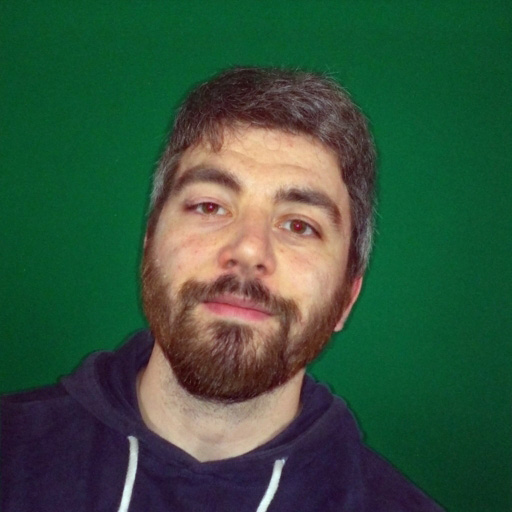}\label{fig:val_p2p}}&
	{\includegraphics[width=0.235\textwidth]{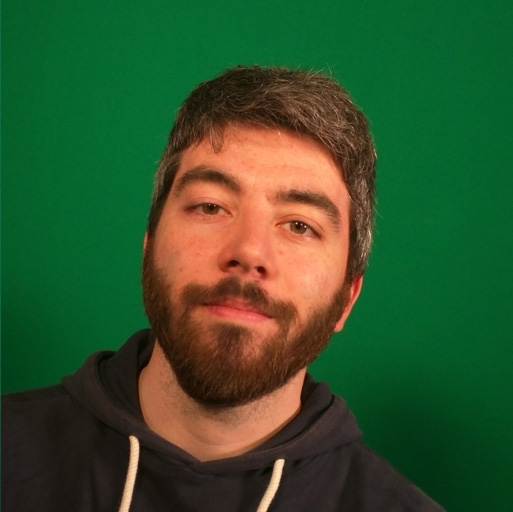}\label{fig:val_p2p_target}}\\
	
	\rotatebox{90}{\ \ \ \ \ \ \ Test}&{\includegraphics[width=0.235\textwidth]{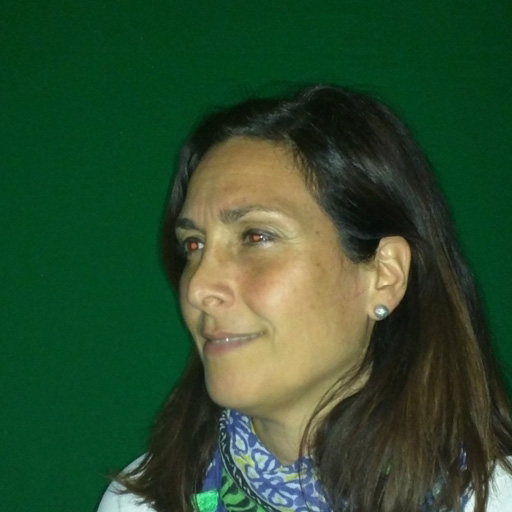}\label{fig:test_p2p_input}}&
	{\includegraphics[width=0.235\textwidth]{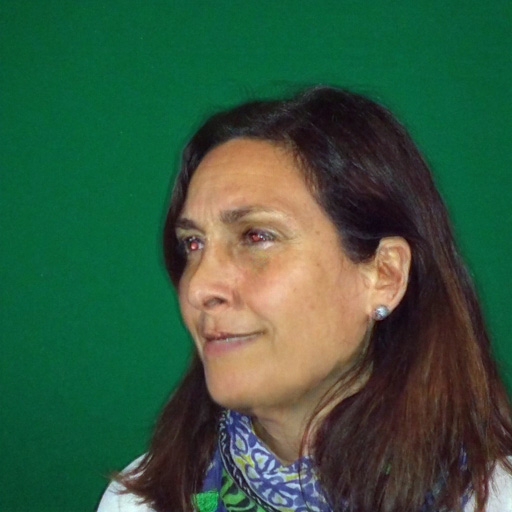}\label{fig:test_p2p_h}}&
	{\includegraphics[width=0.235\textwidth]{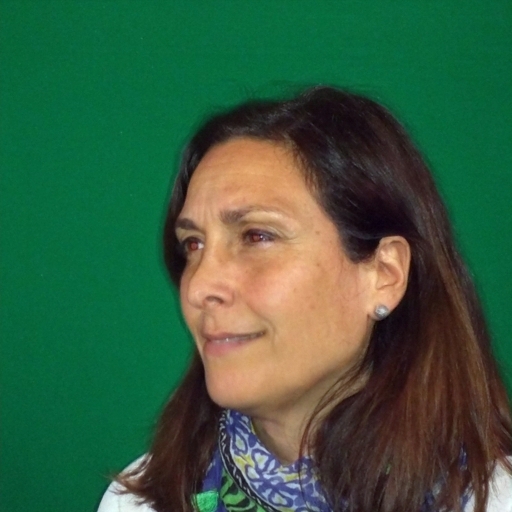}\label{fig:test_p2p}}&
	{\includegraphics[width=0.235\textwidth]{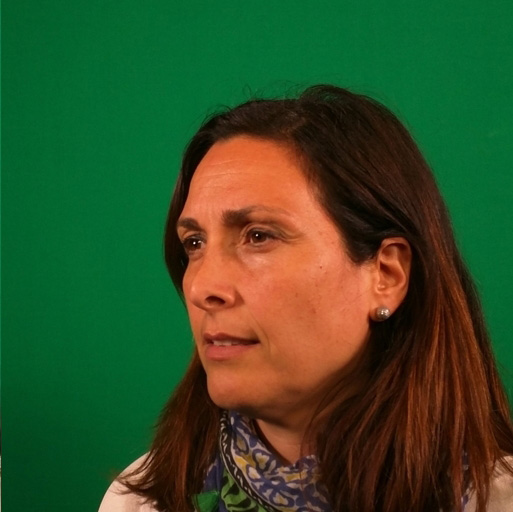}\label{fig:test_p2p_target}}\\
	\end{tabular}
	
\caption{\label{fig:comp_p2p} \newfixme{An example of comparisons between Pix2Pix (with our problem encoding) and our approach. We show one example from the training set, the validation set and the test set, respectively.}}
\end{figure}

\newfixme{
We also compared our approach against Pix2Pix by Isola et al.~\cite{pix2pix2017}. Such work is based on a particular type of Generative Adversarial Network (GAN)~\cite{goodfellow2014generative} in the conditional setting (cGAN)~\cite{mirza2014conditional}. The use of such type of neural network was investigated as a general-purpose solution to image-to-image translation problems. Isola et al. tested their cGAN on several tasks such as photo generation and semantic segmentation. 

In the training details, they explained that the images were randomly jittered through resizing the $256 \times 256$ (original size of images) to $286 \times 286$ and then randomly cropped to come back to $256 \times 256$. Therefore, we trained Pix2Pix cGAN by using the training information provided for the $Day \rightarrow Night$ task, which is performed by the authors. In particular, the network was trained using our dataset; i.e., 4 as batch size and 80 as the number of epochs. We had to increase the number of epochs because 17 epochs, as in the case of the $Day \rightarrow Night$ task proposed by Isola et al., produced low quality results.


The network was trained from scratch by using a Gaussian distribution to initialize the weights with 0 as mean and 0.02 as standard deviation, as suggested by the authors. We performed a mirroring and the random jitter starting from our image resolution $512 \times 512$ and doubling the resizing to $572 \times 572$. 

As the previous comparison (see Section~\ref{HDRNET:COMP}), we performed two experiments. In the first one, we trained Pix2Pix end-to-end using our dataset by feeding the network with input and target images. The network was trained for about 9 hours. Figure~\ref{fig:comp_ee_p2p} shows the comparison with our approach. Although Pix2Pix better approximates the ground truth image colors, it introduces notable artifacts by changing the image content significantly. Many of these artifacts can be seen on eyes and facial features.

In the second experiment, we combined Pix2Pix with our encoding. We trained Pix2Pix with the filtered images as input and the difference between the input and ground truth filtered as a target; see Section~\ref{problem_encoding}. In this case, the network was trained for 9 hours with the same parameters used in the first case. Figure~\ref{fig:comp_p2p} shows some outcomes of this experiment. As the previous comparison, it is possible to notice a dramatic improvement of the results obtained by combining Pix2Pix and our encoding. However, artifacts are still present on the geometry of the image (i.e., facial features and around eyes), even though they are not as strong as in the end-to-end training.

The results of the comparisons are visible in the Table~\ref{tab_comp_p2p}, which reports the SSIM and PSNR values. As the previous comparison, these values are computed on a random sample of 30 images for each subset of the dataset (i.e., training set, validation set, and test set). From this table, it can be seen that our approach obtains higher values of SSIM and PSNR for all subsets.

Finally, this comparison elicits that our approach outperforms Pix2Pix w.r.t the image operator that remove the flash artifacts. Furthermore, Pix2Pix can benefit from our encoding strategy by generating high-quality results compared to end-to-end training.}

\begin{table}[!ht]
  \centering
  \begin{tabular}{|p{2.0cm}|p{2.0cm}|p{2.0cm}|p{2.0cm}|p{2.0cm}|}
    \hline
     & Our SSIM & Pix2Pix SSIM & Our PSNR & Pix2Pix PNSR \\ \hline
    Training                 & \textbf{90\%} & 71\% & \textbf{24.8 dB} & 17.82 dB   \\ \hline
    Validation               & \textbf{83.16\%} & 73.16\% & \textbf{19.72 dB} & 17.64 dB   \\ \hline
    Test                   & \textbf{88.78\%} & 74.25\% & \textbf{20.58 dB} & 17.30 dB   \\ \hline
  \end{tabular}
  \newline\newline
  \caption{\newfixme{Comparisons on samples of the training, validation, and test sets. As before, the SSIM and PSNR values are computed on samples of 30 images each, which were extracted randomly from each dataset.}}\label{tab_comp_p2p}
\end{table}
\newpage
\subsection{Comparisons with Style Transfer}
\begin{figure}[!ht]
\centering
\setlength{\tabcolsep}{1pt}
\begin{tabular}{cccc}
	Input & Ground Truth & Style Transfer & Our Approach\\
	{\includegraphics[width=0.245\textwidth]{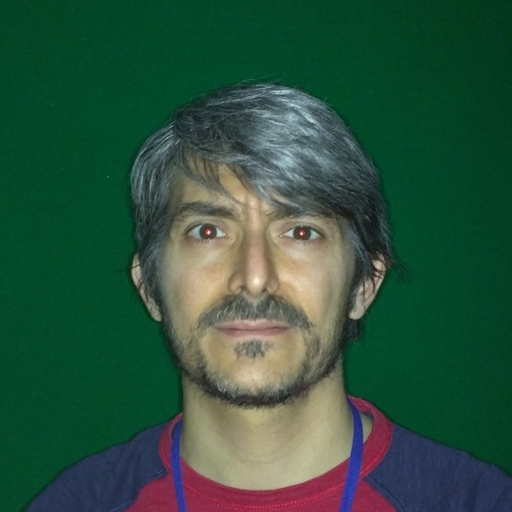}}&
	{\includegraphics[width=0.245\textwidth]{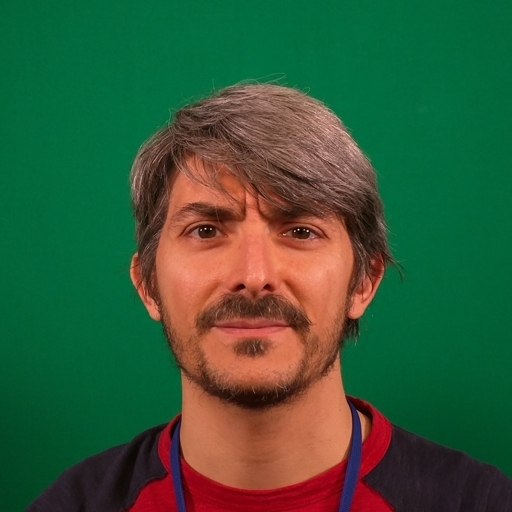}}&
	{\includegraphics[width=0.245\textwidth]{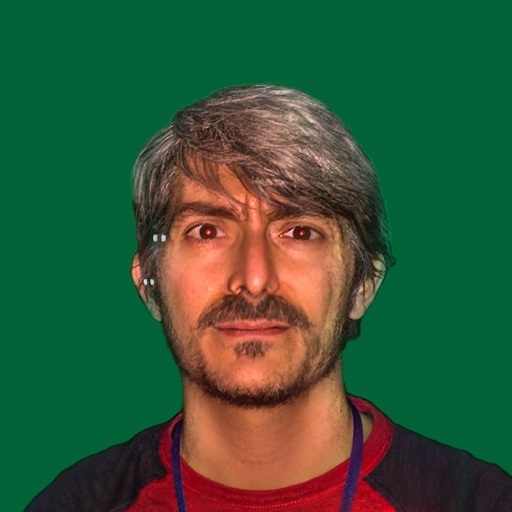}}&
	{\includegraphics[width=0.245\textwidth]{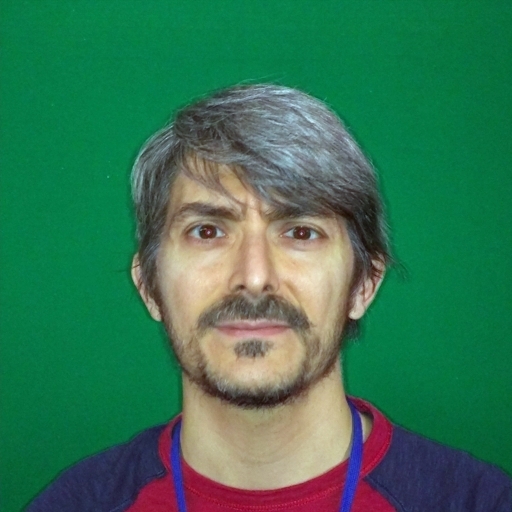}}\\
	{\includegraphics[width=0.245\textwidth]{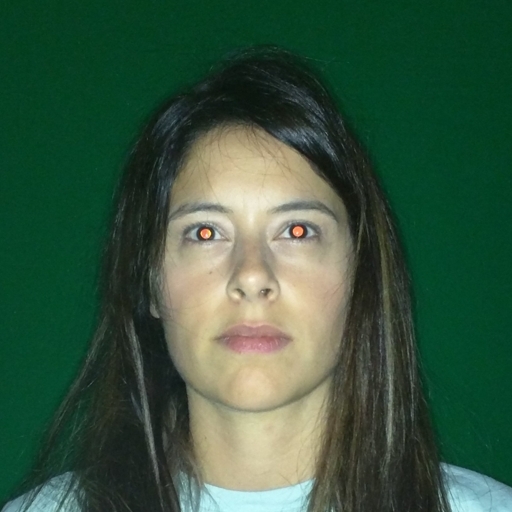}}&
	{\includegraphics[width=0.245\textwidth]{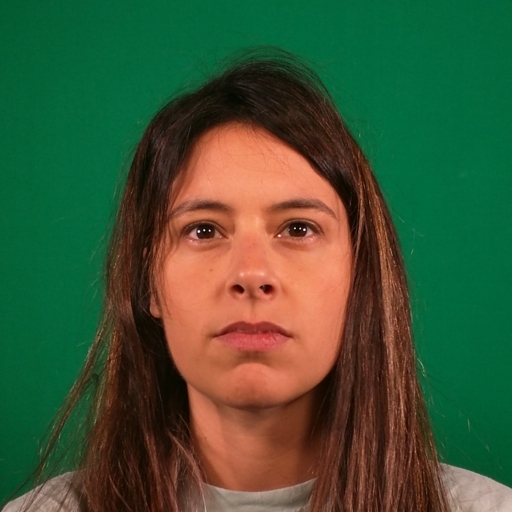}}&
	{\includegraphics[width=0.245\textwidth]{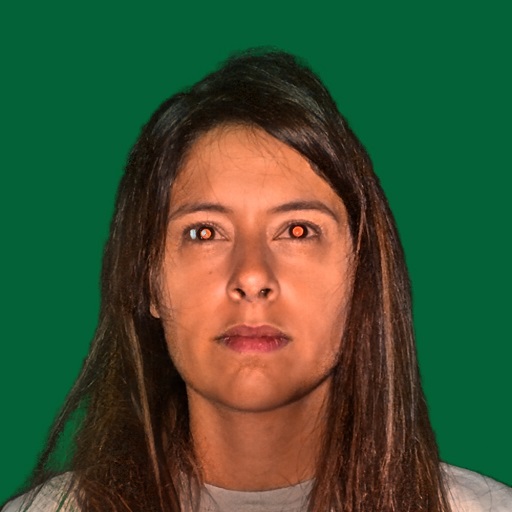}}&
	{\includegraphics[width=0.245\textwidth]{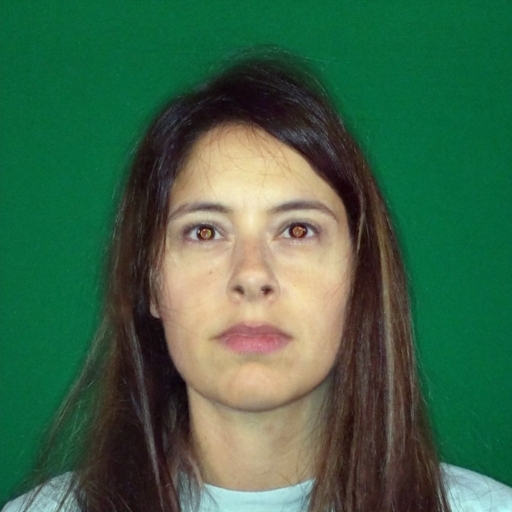}}\\
	{\includegraphics[width=0.245\textwidth]{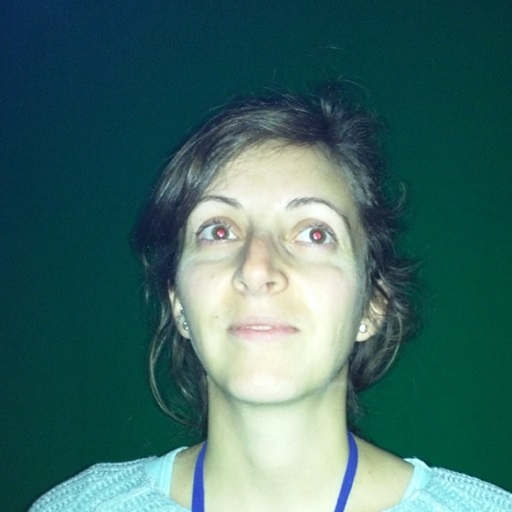}}&
	{\includegraphics[width=0.245\textwidth]{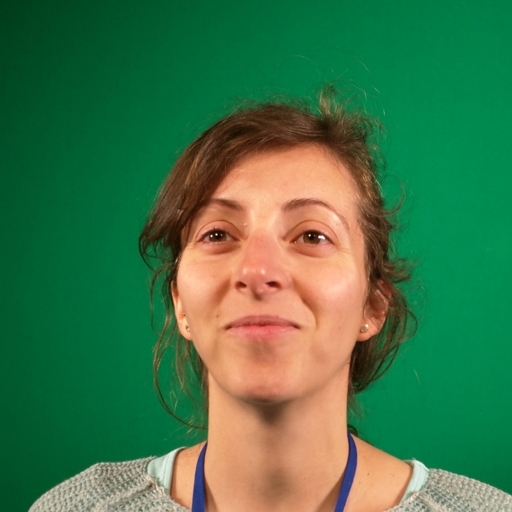}}&
	{\includegraphics[width=0.245\textwidth]{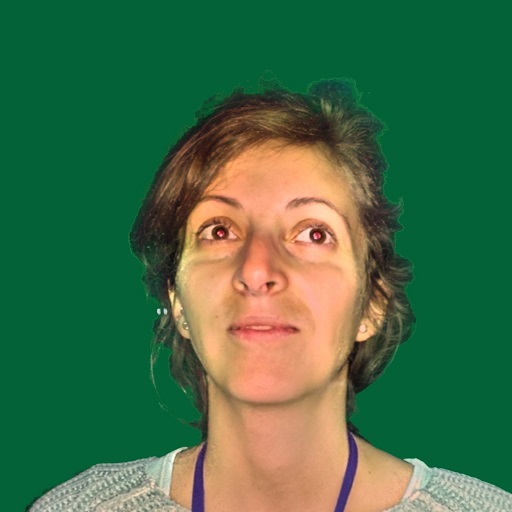}}&
	{\includegraphics[width=0.245\textwidth]{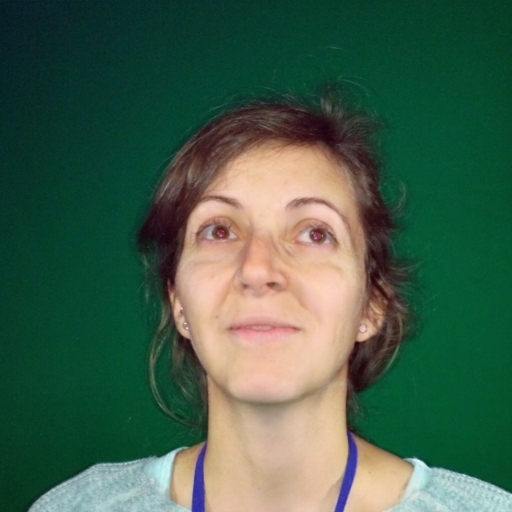}}\\
	{\includegraphics[width=0.245\textwidth]{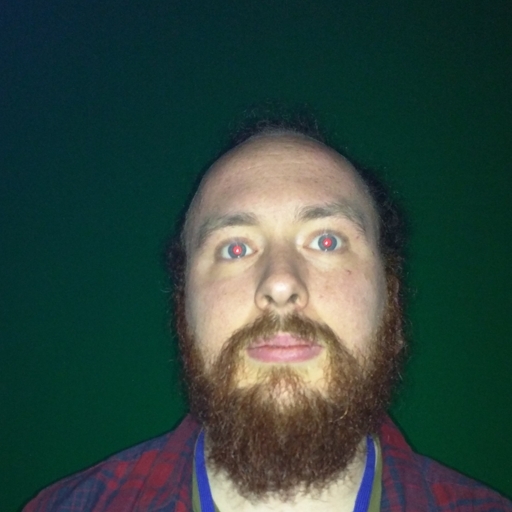}}&
	{\includegraphics[width=0.245\textwidth]{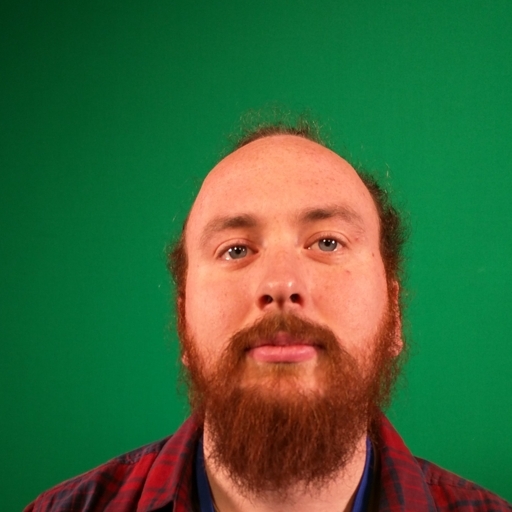}}&
	{\includegraphics[width=0.245\textwidth]{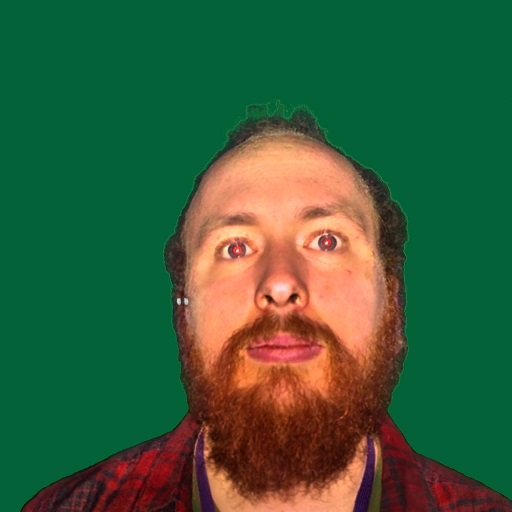}}&
	{\includegraphics[width=0.245\textwidth]{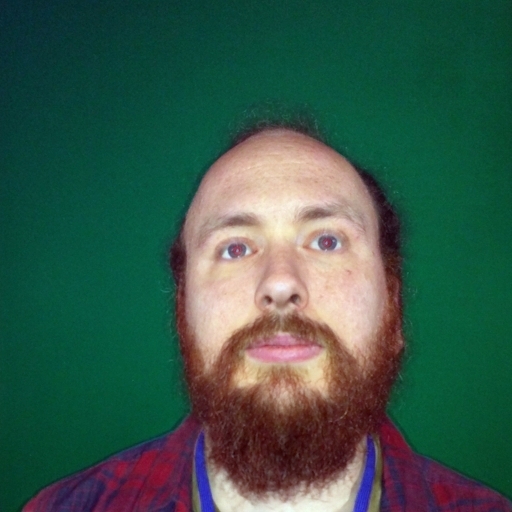}}\\
\end{tabular}
\caption{\label{fig:comp_style} \newfixme{An example of visual comparisons between the Portrait Style transfer~\cite{shih2014style} and our approach. Although there is a good match in colors several artifacts are presents such as flash hard shadows and around eyes.}}
\end{figure}

\newfixme{
As our final comparison, we tested our method against the style transfer method by Shih et al.~\cite{shih2014style}. This method is specifically meant for portraits and based on a multi-scale local transfer approach. In our tests, we used the ground-truth image as target style to be transferred to the input image that is the ideal condition.
Unfortunately, we could not try Shih et al.'s method on a large dataset because the generation of masks and landmarks for the original code is extremely cumbersome (i.e., more than an hour per image). Therefore, we tested on a limited number of images that are displayed on Figure~\ref{fig:comp_style}. As can be seen from Figure~\ref{fig:comp_style}, the method by Shih et al. can transfer colors correctly, but it fails to remove flash artifacts. Moreover, these are enhanced creating unnatural effects especially for eyes and hard shadows and lighting.
}

\section{Limitations}\label{sec:limitations}
Thanks to the tightly bounded domain of the input, our system can provide convincing results even after being trained with a small dataset of 495 input images (prior to data augmentation). However, results would greatly improve if a more diverse training dataset was provided by relaxing some boundaries of the said domain. More specifically: all the subjects were white adults, the uniform-light setting was the same for all images, and all images were acquired with the same device.


\section{Conclusions}\label{sec:conclusion}
We have proposed an unassisted pipeline to turn a smartphone flash selfie into a studio portrait by using a regression model based on supervised learning. We have defined a complete pipeline, starting from data collected by well-defined acquisition parameters, performing pre-processing by a bilateral filter, training the network, and finally, validating the results.

We have made several comparisons using different metrics to validate our approach. Among these, we used the SSIM that places more emphasis on the validity of the results because it measures the similarity between images in a way that is consistent with the perception of the human eye. 
Besides the obvious application of our method for correcting flash selfies, our results allow us to conjecture that a low-quality smartphone flash selfie contains enough information for reconstructing the actual appearance of a human face as one obtained with more uniform lighting.

The most likely future work will be to widen the acquisition domain, both in terms of hardware and illumination settings and in terms of the age and ethnicity of photographed subjects. 
Then, we will build on our method by incorporating state-of-the-art solutions for multiple-face detection, red-eye removal, and background subtraction. Our aim is to deploy a mobile app that can be used by any smartphone user.

On the method itself, we are planning to explore the solution proposed by Larsen et al.~\cite{Larsen16} and to use our network as the generator component of a GAN~\cite{goodfellow2014generative} where the discriminator is trained to recognize if a given  image is a difference between the bilateral filtered version of a flash and no flash images. 
\revision{To improve our CNN encoder-decoder architecture as a future work, we could use ResNet or other models as encoder component~\cite{chaurasia2017linknet} of our CNN as mentioned in recent works~\cite{shvets2018angiodysplasia, iglovikov2018ternausnet}}.


\subsection*{Acknowledgments}
The authors thank NVIDIA's Academic Research Team for providing the Titan Xp card under the Hardware Donation Program. \revision{We wish also to thank Pietro Cignoni for building the hardware driving the flash lighting  and the many colleagues of the Visual Computing Lab and of the whole ISTI-CNR who helped us in building up the photo dataset; in particular Mauro Boni and Giuliano Kraft for providing skilled technical assistance in setting up the light studio.}

\bibliographystyle{unsrt}  
\bibliography{main}

\end{document}